\documentclass[preprint,12pt]{elsarticle}

\usepackage{amssymb}

\usepackage{url}
\usepackage{booktabs}
\usepackage{multirow}
\usepackage{amssymb}
\usepackage{pifont}
\newcommand{\cmark}{\ding{51}}%
\usepackage{amsmath,array,graphicx}
\usepackage{float}
\usepackage{algorithmic}
\usepackage[ruled,linesnumbered]{algorithm2e}
\usepackage[hidelinks]{hyperref}
\usepackage{xcolor}
\usepackage[margin=1in]{geometry}

\usepackage{fancyhdr}

\newcommand{\ha}[1]{\textcolor{black}{#1}}

\begin{document}

\makeatletter
\def\ps@pprintTitle{%
  \let\@oddhead\@empty
  \let\@evenhead\@empty
  \let\@oddfoot\@empty
  \let\@evenfoot\@oddfoot
}
\makeatother

\fancypagestyle{pprintTitle}{%
\lhead{Accepted to the Elsevier Computers \& Security on 20 December 2023} \chead{}\rhead{\scriptsize}
\lfoot{}\cfoot{}\rfoot{}
\renewcommand{\headrulewidth}{0.1pt}
}

\begin{frontmatter}



\title{EvadeDroid: A Practical Evasion Attack on Machine Learning for Black-box Android Malware Detection}


\author[inst1]{Hamid Bostani}
\affiliation[inst1]{organization={Digital Security Group},
            organization={Institute for Computing and Information Science},
            organization={Radboud University},
            city={Nijmegen},            
            country={The Netherlands. E-mail: hamid.bostani@ru.nl (Corresponding Author)}}

\author[inst2]{Veelasha Moonsamy}
\affiliation[inst2]{organization={Horst Goertz Institute for IT Security},
            organization={Ruhr University Bochum},   
            city={Bochum},           
            country={Germany. E-mail: email@veelasha.org}}

\begin{abstract}
Over the last decade, researchers have extensively explored the vulnerabilities of Android malware detectors to adversarial examples through the development of evasion attacks; however, the practicality of these attacks in real-world scenarios remains arguable. The majority of studies have assumed attackers know the details of the target classifiers used for malware detection, while in reality, malicious actors have limited access to the target classifiers. 
This paper introduces \emph{EvadeDroid}, a problem-space adversarial attack designed to effectively evade black-box Android malware detectors in real-world scenarios. 
EvadeDroid constructs a collection of problem-space transformations derived from benign donors that share opcode-level similarity with malware apps by leveraging an \textit{n}-gram-based approach. These transformations are then used to morph malware instances into benign ones via an iterative and incremental manipulation strategy. 
The proposed manipulation technique is a query-efficient optimization algorithm that can find and inject optimal sequences of transformations into malware apps. \ha{Our empirical evaluations, carried out on 1K malware apps, demonstrate the effectiveness of our approach in generating real-world adversarial examples in both soft- and hard-label settings. Our findings reveal that EvadeDroid can effectively deceive diverse malware detectors that utilize different features with various feature types. Specifically, EvadeDroid achieves evasion rates of 80\%-95\% against DREBIN, Sec-SVM, ADE-MA, MaMaDroid, and Opcode-SVM with only 1-9 queries. Furthermore}, we show that the proposed problem-space adversarial attack is able to preserve its stealthiness against five popular commercial antiviruses \ha{with an average of 79\% evasion rate}, thus demonstrating its feasibility in the real world.

\end{abstract}


\end{frontmatter}


\section{Introduction}
\label{sec:introduction}

Machine Learning (ML) continues to show promise in detecting sophisticated and zero-day malicious programs~\cite{b8,b9,b10,b11,b12,b13,b14}. However, despite the effectiveness of ML-based malware detectors, 
these defense strategies are vulnerable to \textit{evasion attacks}~\cite{b15}.
More concretely, attackers aim to deceive ML-based malware classifiers by transforming existing malware into \textit{adversarial examples} (AEs) via a series of manipulations. The proliferation of Android malware~\cite{b16} has extended research into novel evasion attacks to strengthen malware classifiers against AEs~\cite{b17,b18,b19,b20,b21,b22,b23,b24,b25,b26,b27,b28,b29}. However, this endeavor, which also exists for other platforms, such as Windows, poses its own set of challenges, which we elaborate on further below.

The first challenge pertains to the \emph{feature representation} of Android applications (apps). Making a slight modification in the feature representation of a malware app may break its functionality~\cite{b15} as malware features extracted from Android Application Packages (APKs) are usually discrete (e.g., app permissions) instead of continuous (e.g., pixel intensity in a grayscale image). One plausible solution is to manipulate the features extracted from the Android Manifest file~\cite{b17,b20,b24}; however, the practicality of such manipulations in generating executable AEs is questionable for the following reasons. Firstly, modifying features from the Android Manifest (e.g., content providers, intents, etc.) cannot guarantee the executability of the original apps (i.e., malicious payload)~\cite{b25,b33}. Secondly, adding unused features to the Manifest file can be discarded by applying pre-processing techniques~\cite{b26}. Finally, advanced Android malware detectors (e.g.,~\cite{b31,b34}) primarily rely on the semantics of Android apps, which are represented by the Dalvik bytecode rather than the Manifest files~\cite{b27, b24}.

Another challenge is the limitations of \emph{feature mapping} techniques used to convert Android apps from the problem space (i.e., input space) to feature space. These techniques are not reversible, meaning that feature-space perturbations cannot be directly translated into a malicious app~\cite{b26}. To address  \textit{inverse feature-mapping problem}, a common approach is to manipulate real-world malware apps using problem-space transformations that correspond to the features used in ML models. 
By applying these feature-based transformations to Android apps, adversaries can create hazardous evasion attacks~\cite{b26,b27,b28,b29}. However, finding suitable transformations that satisfy problem-space constraints is not straightforward~\cite{b26}: Firstly, certain transformations (e.g.,~\cite{b75,b90}) intended to mimic feature-space perturbations may not result in feasible AEs because they disregard feature dependencies from real-world objects. Additionally, some transformations (e.g.,~\cite{b26,b28}) that meet problem-space constraints for manipulating real objects may introduce undesired or incompatible payloads into malware apps. These types of transformations not only might render the perturbations different from what the attacker expects~\cite{b28} but can also lead to the crashing of adversarial malware apps.

The final challenge revolves around current methods~\cite{b17,b18,b19,b20,b21,b22,b23,b24,b26,b27,b28,b75,b90} that generate AEs based on the specifics of target malware detectors, such as the ML algorithm and feature set. These approaches assume that attackers possess either \textit{Perfect Knowledge} (PK) or \textit{Limited Knowledge} (LK) about the target classifiers. However, in real-world scenarios, adversaries generally have \textit{Zero Knowledge} (ZK) about the target malware detectors, which aligns more closely with reality since antivirus systems operate as black-box engines that are queried~\cite{b84}. Some studies~\cite{b23,b62,zhang2021shadowdroid} have explored semi-black-box settings to generate AEs by leveraging feedback from the target detectors. Nevertheless, these approaches suffer from inefficiency in terms of evasion costs, including the high number of queries required and the extent of manipulation applied to the input sample. Efficient querying is crucial due to the associated costs~\cite{b84} and the risk of detectors blocking suspicious queries. Additionally, minimizing manipulation is desired
as excessive manipulations could impact the malicious functionality of apps~\cite{b19}.

\subsection{Contributions}
In response to the challenges outlined earlier, we propose a comprehensive and generalized evasion attack called \emph{EvadeDroid}, which can bypass black-box Android malware classifiers through a two-step process: (i) \emph{preparation} and (ii) \emph{manipulation}. The first step involves implementing a donor selection technique within EvadeDroid to create an action set comprising a collection of problem-space transformations, i.e., code snippets known as \emph{gadgets}. These gadgets are derived by performing program slicing on benign apps (i.e., donors) that are publicly available. By injecting each gadget into a malware app, specific payloads from a benign donor can be incorporated into the malware app. Our proposed technique utilizes an \emph{n-gram-based similarity} method to identify suitable donors, particularly benign apps that exhibit similarities to malware apps at the opcode level. 
Applying transformations derived from these donors to malware apps can enable them to appear benign or move them towards blind spots of ML classifiers.
This approach aims to achieve the desired outcome of introducing transformations that not only ensure adherence to problem-space constraints (i.e., preserved semantics, robustness to preprocessing, and plausibility~\cite{b26}) but also lead to malware classification errors.

In the manipulation step, EvadeDroid uses an iterative and incremental manipulation strategy to create real-world AEs. This procedure incrementally perturbs malware apps by applying a sequence of transformations gathered in the action set into malware samples over several iterations. We propose a search method to randomly choose suitable transformations and apply them to malware apps. The random search algorithm, which moves malware apps in the problem space, is guided by the labels of manipulated malware apps. These labels are specified by querying the target black-box ML classifier. Our contributions can be summarized as follows:
\begin{itemize}
\item We propose a \textit{black-box evasion} attack that generates real-world Android AEs that adhere to problem-space constraints.
To the best of our knowledge, EvadeDroid is the pioneer study in the Android domain that successfully evades ML-based malware detectors by directly manipulating malware samples without performing feature-space perturbations.

\item We demonstrate that EvadeDroid is a \textit{query-efficient} attack capable of deceiving various black-box ML-based malware detectors through minimal querying. Specifically, our proposed problem-space adversarial attack achieves evasion rates of $89\%$, $85\%$, $86\%$, $95\%$, and $80\%$ against DREBIN~\cite{b30}, Sec-SVM~\cite{b19}, ADE-MA~\cite{b90}, MaMaDroid~\cite{b34}, and Opcode-SVM~\cite{b55}, respectively. This research represents \ha{one of the} pioneering \ha{efforts} in the Android domain, introducing a realistic problem-space attack in a ZK setting.

\item Our proposed attack can operate with either \textit{soft labels} (i.e., confidence scores) or \textit{hard labels} (i.e., classification labels) of malware apps, as specified by the target malware classifiers, to generate AEs. 

\item We assess the practicality of the proposed evasion attack under real-world constraints by evaluating its performance in deceiving popular commercial antivirus products. Specifically, our findings indicate that EvadeDroid can significantly diminish the effectiveness of \ha{five popular} commercial antivirus products, achieving an average evasion rate of approximately \ha{$79\%$}.

\item In the spirit of open science and to allow reproducibility, we have made our code available at {\url{https://github.com/HamidBostani2021/EvadeDroid}}
\end{itemize}
\ha{The rest of the paper is organized as follows: \S\ref{sec:related_work} reviews the most important relevant studies, particularly in the Android domain. In \S\ref{sec:background}, we provide background information on fundamental concepts, specifically ML-based malware detectors, and briefly discuss the practical transformations that can be used for manipulating APKs. \S\ref{sec:proposed_attack} initiates by reviewing the threat model and articulating the problem definition for EvadeDroid. Following this, an illustration of the proposed black-box attack will be presented. We evaluate EvadeDroid's performance in \S\ref{sec:simulation_results}. Limitations and future work, along with a brief conclusion, are presented in \S\ref{sec:limitations_and_future_work} and \S\ref{sec:conclusions}.}
\section{Related Work}
\label{sec:related_work}
In the past few years, several studies have explored AEs in the context of malware, particularly in the Windows domain. For example, Demetrio et al.~\cite{b36} generated AEs in a black-box setting by applying structural and behavioral manipulations. Song et al.~\cite{b41} employed code randomization techniques to generate real-world AEs. \ha{They proposed an adversarial framework guided by reinforcement learning to model the action selection problem as a multi-armed bandit problem.} Sharif et al.~\cite{b79} used binary diversification techniques to evade malware detection. Khormali et al.~\cite{b33} bypassed visualization-based malware detectors by applying padding and sample injection to malware samples. Demetrio et al.~\cite{b82} generated adversarial malware by making small manipulations in the file headers of malware samples. Rosenberg et al.~\cite{b84} presented a black-box attack that perturbs API sequences of malware samples to mislead malware classifiers. 

While evasion attacks have made significant advancements in the Windows domain, their effectiveness in the Android domain may be limited because their manipulations might not be appropriate for altering Android malware apps in a way that can deceive existing Android malware detectors.
Over the last few years, various studies have been performed to generate AEs in the Android ecosystem to anticipate possible evasion attacks. Table~\ref{table:Android-evasions} illustrates the threat models that were considered by researchers. Note that in the categorization of studies under the ZK setting, adversaries should not only lack access to the details of the target model but also have no assumptions (e.g., types of features utilized by detectors) about it. To study feature-space AEs, \ha{Croce et al.~\cite{b62} introduced Sparse-RS, a query-based attack that generated AEs using a random search strategy.} Rathore et al.~\cite{b17} generated AEs by using Reinforcement Learning to mislead Android malware detectors. Chen et al.~\cite{b18,b21} implemented different feature-based attacks (e.g., brute-force attacks) to evaluate their defense strategies. Demontis et al.~\cite{b19} presented a white-box attack to perturb feature vectors of Android malware apps regarding the most important features that impact the malware classification. Liu et al.~\cite{b22} introduced an automated testing framework based on a Genetic Algorithm (GA) to strengthen ML-based malware detectors. Xu et al.~\cite{b23} proposed a semi-black-box attack that perturbs features of Android apps based on the simulated annealing algorithm. The above attacks seem impractical as they do not show how real-world apps can be reconstructed based on feature-space perturbations.

\begin{table}[t]
\footnotesize
\caption{Evasion attacks in ML-based Android Malware Detectors.}
\label{table:Android-evasions}
\begin{center}
\begin{tabular}{l|ccc|cc}
  \toprule
  \multirow{2}{*}{\textbf{Relevant Papers}} 
      & \multicolumn{3}{c|}{\textbf{Attacker's~Knowledge}} 
          & \multicolumn{2}{c}{\textbf{Perturbation~Type}} \\  \cline{2-6}
  & \textbf{PK} & \textbf{LK} & \textbf{ZK} & \textbf{Problem Space} & \textbf{Feature Space} \\  \midrule
    \ha{Xu et al.~\cite{xu2023gendroid}} & & & \ha{\checkmark } & \ha{\checkmark} &  \ha{\checkmark}\\  
      \ha{He et al.~\cite{he2023efficient}} & &  & \ha{\checkmark} & \ha{\checkmark} & \ha{\checkmark}\\  
  Li et al.~\cite{li2023black} & & \checkmark & & \checkmark &  \checkmark\\  
  \ha{Croce et al.~\cite{b62}} & & \ha{\checkmark } & & &  \ha{\checkmark}\\    
  Zhang et al.~\cite{zhang2021shadowdroid} & & \checkmark & & \checkmark & \checkmark \\     
  Rathore et al.~\cite{b17} & \checkmark & \checkmark &  &  & \checkmark \\     
  Chen et al.~\cite{b18} & \checkmark & \checkmark &  &  & \checkmark\\     
  Demontis et al.~\cite{b19} & \checkmark & \checkmark & \checkmark & \checkmark & \checkmark\\      
  Grosse et al.~\cite{b20} & \checkmark &  &  & \checkmark & \checkmark\\  
  Chen et al.~\cite{b21} & \checkmark & \checkmark &  &  & \checkmark\\     
  Liu et al.~\cite{b22} &   & \checkmark &  &  & \checkmark\\    
  Xu et al.~\cite{b23} &  & \checkmark &  &  & \checkmark \\     
  Berger et al.~\cite{b24} & \checkmark & \checkmark &  & \checkmark & \checkmark\\    
  Pierazzi et al.~\cite{b26} & \checkmark &  &  & \checkmark & \checkmark\\     
  Chen et al.~\cite{b27} &  & \checkmark &  & \checkmark & \checkmark\\     
  Cara et al.~\cite{b28} &  & \checkmark &  & \checkmark & \checkmark\\      
  Yang et al.~\cite{b29} &  &  & \checkmark & \checkmark & \checkmark\\   
  Li et al.~\cite{b75} & \checkmark & \checkmark &  & \checkmark & \checkmark\\      
  Li et al.~\cite{b90} & \checkmark & \checkmark &  & \checkmark & \checkmark\\     
  \textbf{EvadeDroid} &  &  & \cmark & \cmark & \\      
  \bottomrule
\end{tabular}
\end{center}
\end{table}

To investigate problem-space manipulations, Grosse et al.~\cite{b20} manipulated the Android Manifest files based on the feature-space perturbations. Berger et al.~\cite{b24} and Li et al.~\cite{b75,b90} used a similar approach; however, they considered both Manifest files and Dalvik bytecodes of Android apps in their modification methods. Zhang et al.~\cite{zhang2021shadowdroid} introduced an adversarial attack called \textit{ShadowDroid} to generate AEs using a substitute model built on permissions and API call features. \ha{Xu et al.~\cite{xu2023gendroid} introduced GenDroid, a query-based attack that employed GA by integrating an evolutionary strategy based on Gaussian Process Regression.} The practicality of these attacks is also questionable because the generated AEs \ha{might not satisfy all the constraints in the problem space}~\cite{b26} (e.g., \ha{plausibility} and robustness to preprocessing). For instance, Li et al.~\cite{b75} reported that 5 out of 10 manipulated apps that were validated could not run successfully. Furthermore, unused features \ha{injected into APKs} by the attacks discussed in\ha{~\cite{b20,b24,b75,b90,zhang2021shadowdroid,xu2023gendroid} not only raise plausibility concerns but also render them susceptible to elimination by preprocessing operator}~\cite{b26}, \ha{especially those features incorporated into} Manifest files.

In addition to the aforementioned studies, some (e.g.,~\cite{b26,b27,b28,b29}) have considered the \textit{inverse feature-mapping problem} when presenting practical AEs in the Android domain. Pierazzi et al.~\cite{b26} proposed a problem-space adversarial attack to generate real-world AEs by applying functionality-preserving transformations to the input malware apps. Chen et al.~\cite{b27} added adversarial perturbations found by a substitute ML model to Android malware apps. Cara et al.~\cite{b28} presented a practical evasion attack by injecting system API calls determined via mimicry attack on APKs. Li et al.~\cite{li2023black} proposed a problem-space attack called \textit{BagAmmo}, targeting function call graph (FCG) based malware detection. 
The main shortcoming of these studies is that the authors assume the adversary to have perfect knowledge\cite{b26} or limited knowledge~\cite{b27,b28} about the target classifiers (e.g., knowing the feature space or accessing the training set), while in real scenarios (e.g., bypassing antivirus engines), an adversary often has zero knowledge about the target malware detectors. For instance, BagAmmo~\cite{li2023black} assumes that the target malware detector is based on FCG, which implies that it has some knowledge about the target model. Note that this assumption may not be applicable in all real-world scenarios, as different malware detectors may employ diverse feature sets.

On the other hand, despite the practicality of~\cite{b26} in attacking white-box malware classifiers, the side-effect features that appear from undesired payloads injected into malware samples may manipulate the feature representations of apps differently from what the attacker expects~\cite{b28}. Furthermore, such attacks may cause the adversarial malware to grow infinitely in size as they do not consider the size constraint of the adversarial manipulations. The attacks presented in~\cite{b27,li2023black} are tailored to the target malware classifiers (i.e., DREBIN~\cite{b30}, and FCG-based detectors such as MaMaDroid~\cite{b34}), which means the authors did not succeed in presenting a generalized evasion technique. Moreover, the attack in~\cite{b28} has some limitations, such as injecting incompatible APIs into Android apps or using incorrect parameters for API calls, which can crash adversarial malware apps.


\ha{To address the aforementioned shortcomings,} Yang et al.~\cite{b29} \ha{proposed} two attacks named the \textit{evolution} and \textit{confusion} attacks, designed to evade target classifiers in a black-box setting. However, their approach lacks details about critical issues (e.g., the feature extraction method) and is impractical because, as reported by the authors, their attacks can easily disrupt the functionality of APKs after a few manipulations. Demontis et al.~\cite{b19} employed an obfuscation tool to bypass Android malware classifiers, but their results indicate a low performance for their method. \ha{He et al.~\cite{he2023efficient} introduced a query-based attack utilizing a perturbation selection tree and an adjustment policy. Nevertheless, the proposed attack is ineffective in hard-label settings, which are crucial for most real-world scenarios. Furthermore, in addition to the questionable plausibility of this attack, its success would be jeopardized by the disputable assumption that perturbations in the attack's malware perturbation set impact the feature values of target malware detectors.}

\ha{EvadeDroid addresses the limitations of existing attacks by thoroughly aiming to meet the practical demands of real-world scenarios, such as hard-label attacking in a fully ZK setting, query efficiency, and satisfaction of all problem-space constraints.} The novelty of our work, compared to the aforementioned studies, lies in the following aspects: (i) EvadeDroid provides adversaries with a general tool to bypass various Android malware detectors, as it is a problem-space evasion attack that operates in a \ha{ZK} setting \ha{without any pre-assumptions about the features and types of features employed by the target malware detectors}~(\S\ref{sec:evasion_costs}). (ii) Unlike other evasion attacks, EvadeDroid directly manipulates Android apps without relying on feature-space perturbations. Its transformations \ha{not only} are independent of the feature space \ha{but also adhere to problem-space constraints~(\S\ref{sec:threat_model})}. (iii) EvadeDroid is simple and easy to implement in real-world scenarios (\S\ref{sec:real_World_scenarios}) \ha{with proper transferability~(\S\ref{sec:transferable_adversarial_examples})}. It is a query-efficient evasion attack that only requires the hard labels of Android apps provided by target black-box malware detectors (e.g., cloud-based antivirus services)~(\S\ref{sec:evasion_costs}). 

\section{Background}
\label{sec:background}
In this section, we present a concise overview of 
\ha{of the fundamental backgrounds relevant to Android evasion attacks.} This encompasses the structure of Android apps, ML-based Android malware detection, \ha{and the adversarial transformations used for generating Android adversarial malware}.
\subsection{Android Application Package (APK)}
\label{appendix:app_structure}
APK is a compressed file format with a \textit{.apk} extension. APKs contain various contents such as Resources and Assets. However, the most crucial contents, particularly for malware detectors, are the Manifest (AndroidManifest.xml) and Dalvik bytecode (classes.dex). The Manifest is an XML file that provides essential information about Android apps, including the package name, permissions, and definitions of Android components. It contains all the metadata required by the Android OS to install and run Android apps. On the other hand, Dalvik bytecode, also known as Dalvik Executable or DEX file, is an executable file that represents the behavior of Android apps.

\textit{Apktool}~\cite{b46} is a popular reverse-engineering tool for the static analysis of Android apps. This reverse-engineering instrument can decompile and recompile Android apps. In the decompilation process, the DEX files of Android apps are compiled into a human-readable code called \textit{smali}. Besides the above tool, \textit{Soot}\cite{b47} and \textit{FlowDroid}~\cite{b48} are two Java-based frameworks that are used for analyzing Android apps. Soot extracts different information from APKs (e.g., API calls) which are then used during static analysis. One of the advantages of Soot for malware detection is its ability to generate call graphs; however, Soot cannot generate accurate call graphs for all apps because of the complexity of the control flow of some APKs. To address this shortcoming, FlowDroid, which is a Soot-based framework, can create precise call graphs based on the app's life cycle. It is worth noting that EvadeDroid uses Apktool, FlowDroid, and Soot in different components of its pipeline to generate adversarial examples.

\subsection{ML-based Android Malware Detection}
Leveraging ML for malware detection has garnered significant interest among cybersecurity researchers in the past decade. ML has demonstrated its potential as an effective solution in static malware analysis, enabling the identification of sophisticated and previously unknown malware through the generalization capabilities of ML algorithms~\cite{b15}. It is important to note that static analysis is a prominent approach for detecting malicious programs, where apps are classified based on their source code (i.e., static features) without execution. This approach offers fast analysis, allowing for the examination of an app's code comprehensively, with minimal resource usage in terms of memory and CPU~\cite{b93}. In order to represent programs for ML algorithms, various types of features are commonly employed in the static analysis, including syntax features (e.g., requested permissions and API calls~\cite{b30,b19,b90}), opcode features (e.g., n-gram opcodes~\cite{b68}), image features (e.g., grayscale representations of bytecodes~\cite{han2015malware}), and semantic features (e.g., function call graphs~\cite{b34}).
\subsection{\ha{Adversarial} Transformations}
In the programming domain, a safe transformation \ha{refers to a problem-space transformation} that maintains the semantic equivalence of the original program while ensuring its excitability. \ha{In the adversarial malware domain, safe transformations, which guarantee preserved-semantics constraint, can become adversarial transformations if they are also plausible and robust to processing (refer to~\ref{appendix:problem_space_constraints} for additional details regarding these constraints). Generally, in} the context of \ha{Android} malware detection, attackers have three types of \ha{adversarial} transformations at their disposal to manipulate malicious \ha{apps}~\cite{b26}: (i) \textit{feature addition}, (ii) \textit{feature removal}, and (iii) \textit{feature modification}. Feature addition involves adding new elements, such as API calls, to the programs, while feature removal entails removing contents like user permissions. Feature modification combines both addition and removal transformations in malware programs. Most studies have primarily focused on feature addition, as removing features from the source code is a complex operation that may cause malware apps to crash. Code transplantation~\cite{b26,b29}, system-predefined transformation~\cite{b28}, and dummy transformation~\cite{b20,b24,b27,b75,b90} are three potential methods for adding features to manipulate Android apps. However, two main issues arise when considering feature additions:

\noindent\textbf{(i) What specific content should be included.} By deriving problem-space transformations from feature-space perturbations, the attacker aims to ensure that the additional contents (e.g., API calls, Activities, etc.) are guaranteed to appear in the feature vector of the manipulated malware \ha{app}~\cite{b26}. Therefore, attackers may either use dummy contents (e.g., functions, classes, etc.)~\cite{b27} or system-predefined contents (e.g., Android system packages)~\cite{b28} for this purpose. \ha{As the plausibility of these transformations is debatable due to the potential lack of complete inconspicuousness,} malicious actors may also make use of content present in already-existing Android apps. The \textit{automated software transplantation} technique~\cite{b35} can then be used to allow attackers to successfully carry out safe transformations. They extract some slices of existing bytecodes from benign apps (i.e., donor) during the \textit{organ harvesting} phase, and the collected payloads are injected into malware apps in the \textit{organ transplantation} phase.

\noindent\textbf{(ii) Where contents should be injected.} New contents must preserve the semantics of malware samples; therefore, they should be injected into areas that cannot be executed \ha{during runtime}. For example, new contents can be added after \texttt{RETURN} instructions~\cite{b19} or inside an \texttt{IF} statement that is always false~\cite{b26}. However, \ha{these injected contents are not robust to preprocessing if} static analysis can discard unreachable code. One creative idea to add unreachable code that is undetectable is the use of \textit{opaque predicates}~\cite{b49}. In this approach, new contents are injected inside an \texttt{IF} statement where its outcome can only be determined at runtime~\cite{b26}.

\section{Proposed Attack}
\label{sec:proposed_attack}
Here we first review the threat model and the problem definition of EvadeDroid. Subsequently, we will offer an illustration of the proposed attack.
\subsection{Threat Model}
\label{sec:threat_model}
\noindent\textbf{Adversarial Goal.} The purpose of EvadeDroid is to manipulate Android malware samples in order to deceive static ML-based Android malware detectors. The proposed attack is an untargeted attack~\cite{b63} \ha{ designed to mislead binary classifiers utilized in Android malware detection, causing Android malware apps to be misclassified.} In other words, EvadeDroid's objective is to trick malware classifiers into classifying malware samples as benign.

\noindent\textbf{Adversarial Knowledge.} 
The proposed evasion attack has black-box access to the target malware classifier. Therefore, EvadeDroid does not have knowledge of the training data $D$, the feature set $X$, or the classification model $f$ (i.e., the classification algorithm and its hyperparameters). The attacker can only obtain the classification results (e.g., hard labels or soft labels) by querying the target malware classifier. 

\noindent\textbf{Adversarial Capabilities.} EvadeDroid is designed to deceive black-box Android malware classifiers during their prediction phase. Our attack manipulates an Android malware \ha{app} by applying a set of safe transformations, known as Android gadgets \ha{(i.e., slices of the benign apps’ bytecode)}, which are optimized through interactions with the black-box target classifier. 
\ha{To ensure adherence to problem-space constraints,} EvadeDroid leverages a tool, developed by the authors~\cite{b26}, \ha{for extracting and injecting gadgets}. 
\ha{Furthermore,} in order to avoid major disruptions to apps, the manipulation process of a malware app is conducted gradually, making it resemble benign apps. This is achieved by injecting a minimal number of gadgets extracted from benign apps into the malware app, \ha{and the process continues until the malware app is misclassified or reaches the predefined evasion cost}. In addition to the problem-space constraints discussed in previous research~\cite{b26}, EvadeDroid must also adhere to two additional constraints highlighting the significance of minimizing evasion costs:
\begin{itemize}
    \item \textbf{\ha{N}umber of queries.} EvadeDroid is a decision-based adversarial attack that aims to generate AEs while minimizing the number of queries, thus reducing the associated costs~\cite{b84}.
    \item \textbf{\ha{Size of} adversarial payloads.} In order to generate executable and visually inconspicuous AEs, such as those with minimal file size~\cite{b36}, EvadeDroid aims to minimize the size of injected adversarial payloads.
\end{itemize}

\ha{It is worth mentioning,} each gadget consists of an organ, which represents a slice of program functionality, an entry point to the organ, and a vein, which represents an execution path that leads to the entry point~\cite{b26}. EvadeDroid extracts gadgets from benign apps by identifying entry points, which are typically API calls, through string analysis. The proposed attack assumes that the benign apps used for gadget extraction are not obfuscated, particularly in terms of their API calls. This is because EvadeDroid relies on string analysis to identify entry points, which limits its ability to extract gadgets from obfuscated apps. \ha{The gadget injection is considered successful when both the classification loss value of the manipulated app increases and the injected adversarial payload conforms to the predefined size of the adversarial payload.} Additionally, the injected gadgets are placed within the block of an obfuscated condition statement that is always evaluated as \texttt{False} during runtime and cannot be resolved during design time.

\noindent\textbf{Defender's Capabilities.} 
In this study, we assume that the target ML models do not employ adaptive defenses that are aware of the operations performed by EvadeDroid due to disclosing detectors' vulnerability to EvadeDroid. Specifically, these target models are unable to enhance their resilience by incorporating AEs generated by EvadeDroid during adversarial training. Furthermore, they lack the capability to detect and block queries from EvadeDroid if they become suspicious of its origin. Importantly, our analysis suggests that EvadeDroid can still be effective even if we relax the second assumption regarding the defender's capabilities. This is supported by empirical evidence demonstrating that our attack often requires only a minimal number of queries to generate AEs.

\subsection{Problem Definition} 
Suppose $\phi: Z \rightarrow X\subset \mathbb{R}^n$ is a feature mapping that encodes an input object $z \in Z$ to a feature vector $x \in X$ with dimension $n$. We denote this as $\phi(Z) = X$. Here, $Z$ represents the input space of Android applications, and $X$ represents the feature space of the app's feature vectors. Furthermore, let $f:X \rightarrow \mathbb{R}^2$ and $g:X \times Y \rightarrow \mathbb{R}$ denote a malware classifier and its discriminant function, respectively. The function $f$ assigns an Android app $z \in Z$ to a class $f(\phi(z)) = \arg\max_{y=0,1} g_y(\phi(z))$, where $y=1$ indicates that $z$ is a malware sample and vice versa. The confidence score (soft label) for classifying $z$ into class $y$ is denoted as $g_y(\phi(z))$. Let $T:Z \overset{\delta \subseteq \Delta}{\longrightarrow} Z$ be a transformation function, denoted as $T_{\delta \subseteq \Delta}(z) = z'$ or simply $T_{\delta}(z) = z'$, which transforms $z \in Z$ to $z' \in Z$ by applying a sequence of transformations $\delta \subseteq \Delta$ such that $z$ and $z'$ have the same functionality. Here, $\Delta = {\delta_1, \delta_2, ..., \delta_n}$ represents an action set consisting of safe manipulations (transformations). Each $\delta_i \in \Delta$ can independently preserve the functionality of a malware sample when applied.

In this study, the objective of the proposed evasion attack is to generate an adversarial example $z^* \in Z$ for a given malware app $z \in Z$ by applying a minimal sequence of transformations $\delta \subseteq \Delta$ to the app, using at most $Q$ queries, while ensuring that the amount of injected adversarial payloads is equal to or lower than $\alpha$. This can be formulated as the following optimization problem:

\begin{equation} \label{eq_optimization1}
\begin{aligned}
\min_{\delta \subseteq \Delta} \quad & |\delta|\\
\textrm{s.t.} \quad & f(\phi(T_\delta (z))) \neq f(\phi(z))\\
  & q \leq Q    \\
  & c(T_\delta (z) , z) \leq \alpha\\
\end{aligned}
\end{equation}

\noindent where $|\delta|$ denotes the cardinality of $\delta$. Additionally, $Q$ and $\alpha$ represent the evasion cost constraints of EvadeDroid, indicating the maximum query budget and the maximum size of adversarial payloads, respectively. The size of adversarial payloads refers to the relative increase in the size of a malware sample after applying $\delta$, and it is measured using the following payload-size cost function:

\begin{equation} \label{eq_relative_size}
c(T_\delta(z) , z) = \frac{[T_\delta(z)] - [z]}{[z]}\times 100
\end{equation}

\noindent where $[.]$ represents the size of an APK. Equation~(\ref{eq_optimization1}) can be translated into the following optimization problem to find an optimal subset of transformations in the action set:

\begin{equation}\label{eq_optimization2}
\begin{aligned}
\underset{\delta \subseteq \Delta}{\arg\max} \quad & g_{y=0}(\phi(T_\delta (z)))\\
\textrm{s.t.} \quad & q \leq Q\\
  & c(T_\delta (z) , z) \leq \alpha\\
\end{aligned}
\end{equation}

\ha{Equation~(\ref{eq_optimization2}) outlines our objective to identify an optimal subset of problem-space transformations $\delta$ within the action set $\Delta$ that leads to misclassification. Specifically, the optimization aims to enhance the confidence score of classifiers in classifying $\phi(T_\delta (z))$, the feature representation of $z$ modified by applying $\delta$, towards the benign class indicated by $0$. Note that the optimization solver is tasked with identifying the optimal $\delta$ with a maximum of $Q$ queries, given that the adversarial payloads do not alter the size of $z$ beyond $\alpha$.}

\subsection{Methodology}
The primary goal of EvadeDroid is to transform a malware app into an adversarial app in such a way that it retains its malicious behavior but is no longer classified as malware by ML-based malware detectors. This is achieved through an iterative and incremental algorithm employed in the proposed attack, \ha{which aims to disguise malware APKs as benign ones.} The attack algorithm \ha{generates} real-world AEs from malware apps \ha{using \emph{problem-space transformations} that satisfy problem-space constraints. These transformations are extracted from benign apps in the wild, which are similar to malware apps using an \emph{n-gram}-based similarity}. 
In this approach, a \emph{random search} algorithm is used to optimize the manipulations of apps. Each malware app undergoes incremental manipulation during the optimization process, where a sequence of transformations is applied in different iterations. \ha{Before delving into the details of the methodology, we offer a brief overview of n-grams and random search.}

\noindent\ha{\textbf{\emph{n}-Grams}\label{appendix:n_gram}
are contiguous overlapping sub-strings} of items (e.g., letters or opcodes) with a length of $n$ from \ha{the given samples} (e.g., texts or programs). This technique captures the frequencies or existence of a unique sequence of items with a length of $n$ in a given sample. In the area of malware detection, several studies have used n-grams to extract features from malware samples~\cite{b50, b51, b52, b53, b54}. These features can be either byte sequences extracted from binary content or opcodes extracted from source codes. \ha{\emph{n}-Grams} opcode analysis is one of the static analysis approaches for detecting Android malware that has been investigated in various related works~\cite{b55, b56, b57, b58, b59}. To conduct such an analysis, the DEX file of an APK is disassembled into smali files. Each smali file corresponds to a specific class in the source code of the APK that contains variables, functions, etc. \ha{\emph{n}-Grams} are extracted from the opcode sequences that appear in different functions of the smali files.

\noindent\ha{\textbf{Random Search}
\label{appendix:random_search}
(RS)~\cite{b61} is a simple yet highly exploratory search strategy that is used in some optimization problems to find an optimal solution.} It relies entirely on randomness, which means RS does not require any assumptions about the details of the objective function or transfer knowledge (e.g., the last obtained solution) from one iteration to another. In the general RS algorithm, the sampling distribution $S$ and the initial candidate solution $x^{(0)}$ are defined based on the feasible solutions of the optimization problem. Then, in each iteration $t$, a solution $x^{(t)}$ is randomly generated from $S$ and evaluated using an objective function regarding $x^{(t-1)}$. This process continues through different iterations until the best solution is found or the termination conditions are met. \ha{It's noteworthy that RS can be a search strategy with high query efficiency in generating AEs}~\cite{b62}. 

\begin{figure}[t]
    \centering
    \includegraphics[width=7cm]{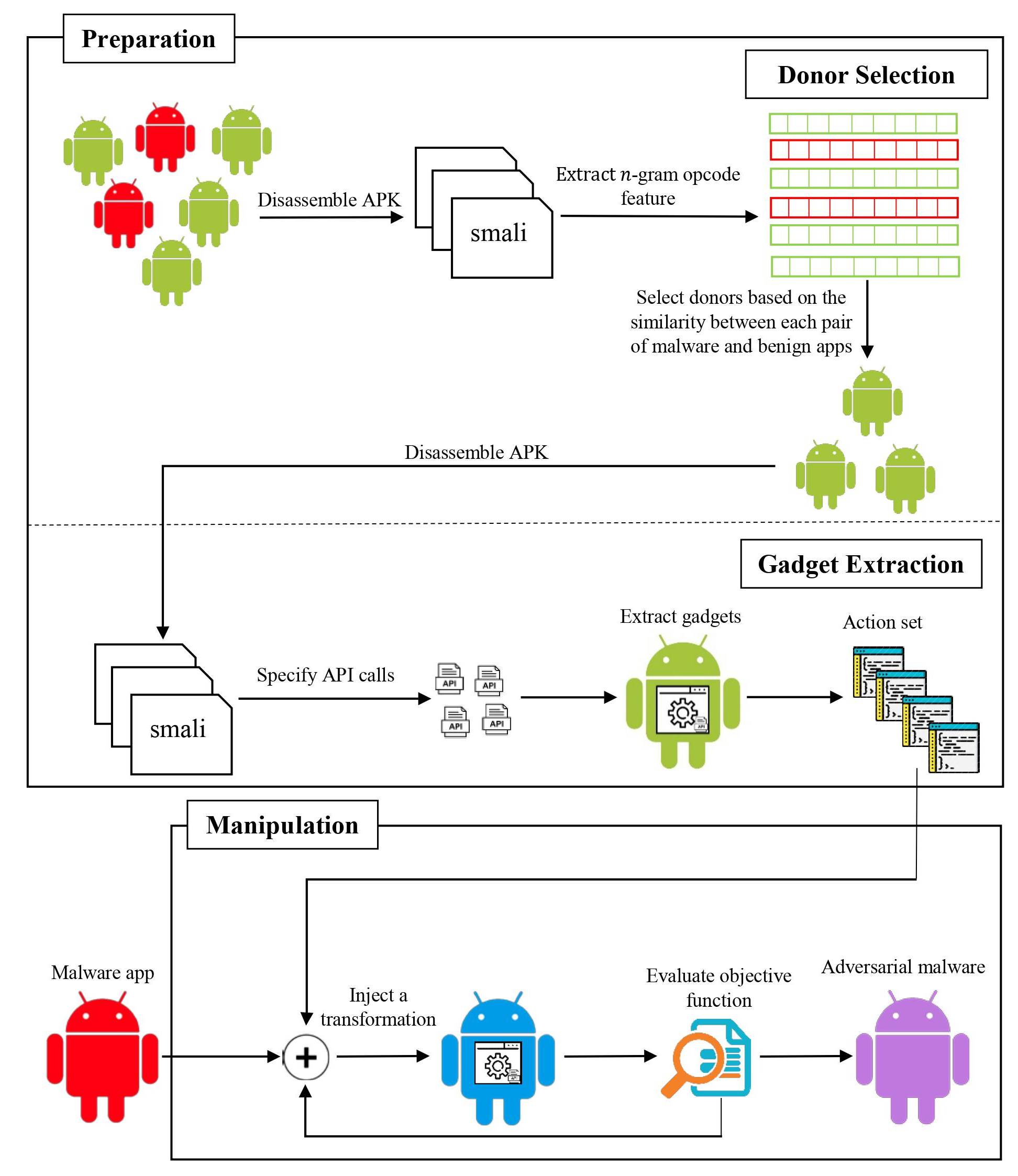}
    \caption{Overview of EvadeDroid's pipeline.}
    \label{fig:block_diagram}
\end{figure}

The workflow of the attack pipeline is illustrated in Figure~\ref{fig:block_diagram}, which consists of two phases: (i) preparation and (ii) manipulation.

\subsubsection{\textbf{Preparation}} 
\label{section:preparation}
The primary objective of this step is to construct an action set comprising a collection of safe transformations that can directly manipulate Android applications. Each transformation in the action set should be capable of altering APKs without causing crashes while preserving their functionality. In this study, program slicing~\cite{b64}, implemented in~\cite{b26}, is utilized to extract the gadgets that make up the transformations collected in the action set. During the preparation step, two important considerations are determining appropriate donors and identifying suitable gadgets. Employing effective gadgets enables the modification of a set of features that can alter the classifier's decision. EvadeDroid achieves this by executing the following two sequential steps:\newline

\noindent\textbf{a) Donor selection.} EvadeDroid selects donors from a pool of benign apps in order to mimic malware instances as benign ones. While it is possible to extract gadgets from any available benign app, collecting transformations from a large corpus of apps is computationally expensive due to the complexity of the program-slicing technique used for organ harvesting. Additionally, identifying potential donors resembling malware apps can lead to obtaining transformations that facilitate disguising malware apps as benign. This is because malware apps that share similarities with benign ones may require fewer transformations to become AEs.
In this study, EvadeDroid adopts a strategy of limiting the number of donors, i.e., choosing donors from the pool of benign apps that resemble malware apps. Our empirical results demonstrate that utilizing transformations from such benign apps accelerates the process of converting malware apps into benign ones, resulting in a reduced number of queries and transformations required for manipulation (refer to~\ref{appendix:donors_evaluation} for more details).

\begin{figure}[t]
    \centering
    \includegraphics[width=6cm]{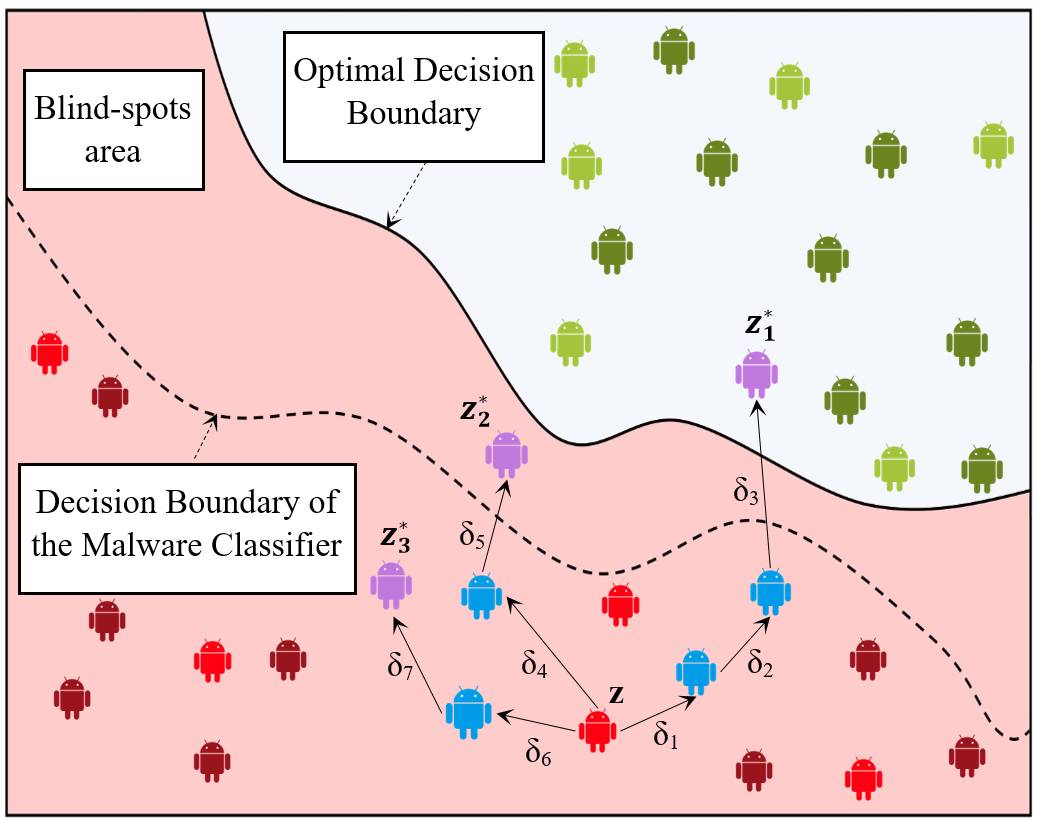}
    \caption{The functionality of EvadeDroid in generating real-world adversarial malware apps. The dark red and dark green samples are, respectively, the inaccessible malware and benign samples that have been used for training the malware classifier. Light red and light green samples represent, respectively, accessible malware and benign samples in the wild. The blue and purple samples are manipulated malware apps and AEs, respectively.}
    \label{fig:conceptual_model}
\end{figure}

More specifically, by utilizing the extracted gadgets from these donors, EvadeDroid can generate effective adversarial perturbations by considering both feature and learning vulnerabilities~\cite{b65,b66}. Figure~\ref{fig:conceptual_model} provides a conceptual representation of EvadeDroid's performance in evading the target classifier. As depicted in Fig.~\ref{fig:conceptual_model}, incorporating segments of benign apps that resemble malware apps can either make malware apps look benign ($T_\delta (z)=z_1^*$ where $\delta=\{\delta_1,\delta_2,\delta_3\}$), or shift them towards the blind spots of the target classifier (e.g., $T_\delta (z)=z_2^*$ where $\delta=\{\delta_4,\delta_5\}$). Note that some sequences of transformations may fail to generate successful AEs (e.g., $\{\delta_6,\delta_7\}$). In this work, we employ an $n$-gram-based opcode technique to assess the similarities between malware and benign samples. Extracting $n$-gram opcode features enables automated feature extraction from raw bytecodes, allowing EvadeDroid to measure the similarity between real objects without requiring knowledge of the feature vector of Android apps in the feature space of the target black-box malware classifiers. We extract $n$-grams following typical approaches found in the literature (e.g.,~\cite{b67,b68}), but with a focus on opcode types rather than the opcodes themselves. The $n$-gram opcode feature extraction utilized in this study involves the following main steps:

\begin{enumerate}
    \item Disassemble Android application's DEX files into smali files using Apktool.
    \item Discard operands and extract $n$-grams from the types of all opcode sequences in each smali file belonging to the app. For example, consider a sequence of opcodes in a smali file: \textit{I: if-eq\space\space\space M: move\space\space\space G: goto\space\space\space I: if-ne\space\space\space M: move-exception\space\space\space G: goto/16\space\space\space M: move-result}. In this case, we have $7$ opcodes with $3$ types (i.e., $I, M, G$). Note \textit{IM, MG, GI, GM} are all unique 2-grams that appeared in the given sequence.
    \item Map the extracted feature sets to a feature space $H$ by aggregating all observable $n$-grams from all APKs.
    \item Create a feature vector $h \in H$ for each app, where each element of $h$ indicates the presence or absence of a specific $n$-gram in the app.
\end{enumerate}

Suppose $M$ and $B$ represent the sets of malware and benign apps, respectively, available to EvadeDroid. The similarity between each pair of a malware app $m_i \in M$ and a benign app $b_j \in B$ is determined by measuring the containment~\cite{b67,b68} of $b_j$ in $m_i$ using the following approach:

\begin{equation}
\label{eq_similarity_containment}
\sigma(m_i,b_j)=\frac{|v(m_i) \cap v(b_j)|} {|v(b_j)|}
\end{equation}

\noindent where $v(m_i)$ and $v(b_j)$ represent the sets of features with values of $1$ in $h_{m_i}$ and $h_{b_j}$, respectively, and $|.|$ denotes the number of features. \ha{Specifically, ${|v(m_i) \cap v(b_j)|}$ denotes the number of common features between $m_i$ and $b_j$.}
It is worth emphasizing that most Android malware apps are created using repackaging techniques, where attackers disguise malicious payloads in legitimate apps~\cite{b70}. Therefore, we consider the containment of benign samples in malware samples to determine the similarities between each pair of malware and benign samples. To identify suitable donors, we calculate a weight for each benign app $b_i \in B$ according to equation~(\ref{eq_similarity_containment}):

\begin{equation}
\label{eq_similarity_weight}
w_{b_j}=\frac{\sum_{\forall m_i \in M} \sigma(m_i,b_j)} {|M|}
\end{equation}

\noindent where $|M|$ represents the number of malware apps. We then sort the benign apps in descending order based on their corresponding weights. Finally, we select the top-$k$ benign apps as suitable donors for gadget extraction. \ha{Note that $w_{b_j}$ reflects how closely $b_j$ aligns with the distribution of malware apps, offering a measure of its resemblance to the characteristics of malware.}\newline

\noindent\textbf{b) Gadget extraction.} 
We collect gadgets based on the desired functionality we aim to extract from donors. EvadeDroid intends to simulate malware samples to benign ones from the perspective of static analysis; therefore, the payloads responsible for the key semantics of donors are proper candidates for extraction. To access the semantics of Android applications, EvadeDroid extracts the payloads containing API calls (i.e., the code snippet encompassing an API call and all its associations) since API calls represent the main semantics of apps~\cite{b71,b72}. An API call is an appropriate point in the bytecode of an APK because the snippets encompassing the API calls are related to one of the app semantics. In sum, gadget extraction from donors consists of the following main steps:




\begin{enumerate}
    \item Disassemble DEX files of donors into smali files by using Apktool.
    \item Perform string analysis on each app to identify all API calls in its smali files.
    \item Extract the gadgets associated with the collected API calls from each app.
\end{enumerate}

Ultimately, the action set $\Delta$ is formed by taking the union of the extracted gadgets.

\subsubsection{\textbf{Manipulation}}
We employ Random Search (RS) as a simple black-box optimization method to solve equation~(\ref{eq_optimization2}). Specifically, for each malware sample $z$, EvadeDroid utilizes RS to find an optimal subset of transformations $\delta$ in order to generate an adversarial example $z^*$. RS offers a significant advantage in terms of query reduction compared to other heuristic optimization algorithms, such as Genetic Algorithms (GAs). This is because RS only requires one query in each iteration to evaluate the current solution. Algorithm~\ref{alg_generate_adm} outlines the key steps of the manipulation component in the proposed problem-space evasion attack. As depicted in Algorithm~\ref{alg_generate_adm}, the RS method randomly selects a transformation $\lambda$ from the action set $\Delta$ to generate $z^*$ for $z$. Subsequently, based on the adversarial payload size $\alpha$, the algorithm applies $\lambda$ to $z$ only if it can improve the objective function $L$ defined in equation~(\ref{eq_optimization2}), which corresponds to the discriminant function of the target classifier for $y=0$.
\begin{algorithm}[t]
\SetAlgoLined
\KwIn{$z$, the original malware sample; $\Delta$, the action set; $L$, the objective function; $\phi$, the feature mapping function; $c$, the payload-size cost function; $Q$, the query budget; $\alpha$, the allowed adversarial payload size.}
\KwOut{ $z^*$, an adversarial example; $\delta$, an optimal transformations.}
 $q \leftarrow 1$ \;
 $z^* \leftarrow z$\;
 $L_{best} \leftarrow $-$\infty$\;
 $\delta \leftarrow $ \O \;
 \While{$q \leq Q$ {\bf and} $z^*$ is classified as a malware}{
  $\lambda \leftarrow$ Select a transformation randomly from $\Delta$ \textbackslash $\delta$\;
  $z^\prime \leftarrow T_{\lambda}(z^*)$\;
  $l = L(\phi(z^\prime))$\;
  \If{$c(z,z^\prime) \leq \alpha$}{
      \If{$ L_{best}\leq l$}{
       $L_{best} \leftarrow l $\;
       $z^* \leftarrow z^\prime$\;
       $\delta \leftarrow \delta \cup \lambda$
       }
   }
 }
 \Return $z^*$, $\delta$
 \caption{Generating a real-world adversarial example.}\label{alg_generate_adm}
\end{algorithm}

\noindent\textbf{Hard-label Setting.} 
In Algorithm~\ref{alg_generate_adm}, we assume that our attack has access to the soft label of the target classifier. This means that EvadeDroid can obtain the confidence score provided by the black-box classification model when making queries. However, in real-world scenarios, such as antivirus systems, the target classifier may only provide hard labels (i.e., classification labels) for Android apps. In this study, we consider two approaches, namely optimal and non-optimal hard-label attacks, to address this challenge. In the optimal hard-label attack, the adversary aims to generate AEs by applying minimal transformations. To achieve this, EvadeDroid modifies the objective function of the proposed RS algorithm (i.e., equation~(\ref{eq_optimization2})) by maximizing the following objective function, while considering the evasion cost:


\begin{equation}\label{eq_optimization_hard_label}
\begin{aligned}
\ha{\underset{\delta \subseteq \Delta}{\arg\max} \quad} & \ha{s(T_\delta (z))}\\
\ha{\textrm{s.t.}} \quad & \ha{q \leq Q}\\
  & \ha{c(T_\delta (z) , z) \leq \alpha}\\
\end{aligned}
\end{equation}

\noindent \ha{where $Q$ (i.e., number of queries) and $\alpha$ (i.e., size of adversarial payloads) represent evasion cost budgets, and $c$ denotes the payload-size cost function (equation~(\ref{eq_relative_size})). Moreover,} $s$ is the following similarity function:

\begin{equation}
\label{eq_similarity}
s(a)=\max_{\forall b \in B} \frac{|v(a) \cap v(b)|} {\lVert h_a-h_b \rVert_1}
\end{equation}

\noindent where $B$ represents all available benign apps in the wild. \ha{$v(a)$ and $v(b)$ represent the sets of features with values of $1$ in $h_{a}$ (i.e., the feature vector of $a$) and $h_{b}$ (i.e., the feature vector of $b$), respectively, and $|.|$ denotes the number of features. } Furthermore, $\lVert h_a-h_b \rVert_1$ denotes the sum of the absolute differences (i.e., $l_1$-norm) between the opcode-based feature vectors of $a$ and $b$. \ha{The $l_1$-norm enhances the accuracy of our similarity measurement, particularly in scenarios where the number of common features between various pairs of malware samples and benign samples is the same, aiding EvadeDroid in identifying the maximum similarity.} Note that equation~(\ref{eq_similarity}) aims to measure the similarity between two apps based on not only a large set of common features but also a small distance. The underlying idea behind the introduced objective function is rooted in our primary approach to misleading malware classifiers. In other words, a transformation can be applied to a malware app if it maintains or increases the maximum similarity between the malware app and available benign apps.

\begin{figure}[t]
    \centering
    \includegraphics[width=13cm]{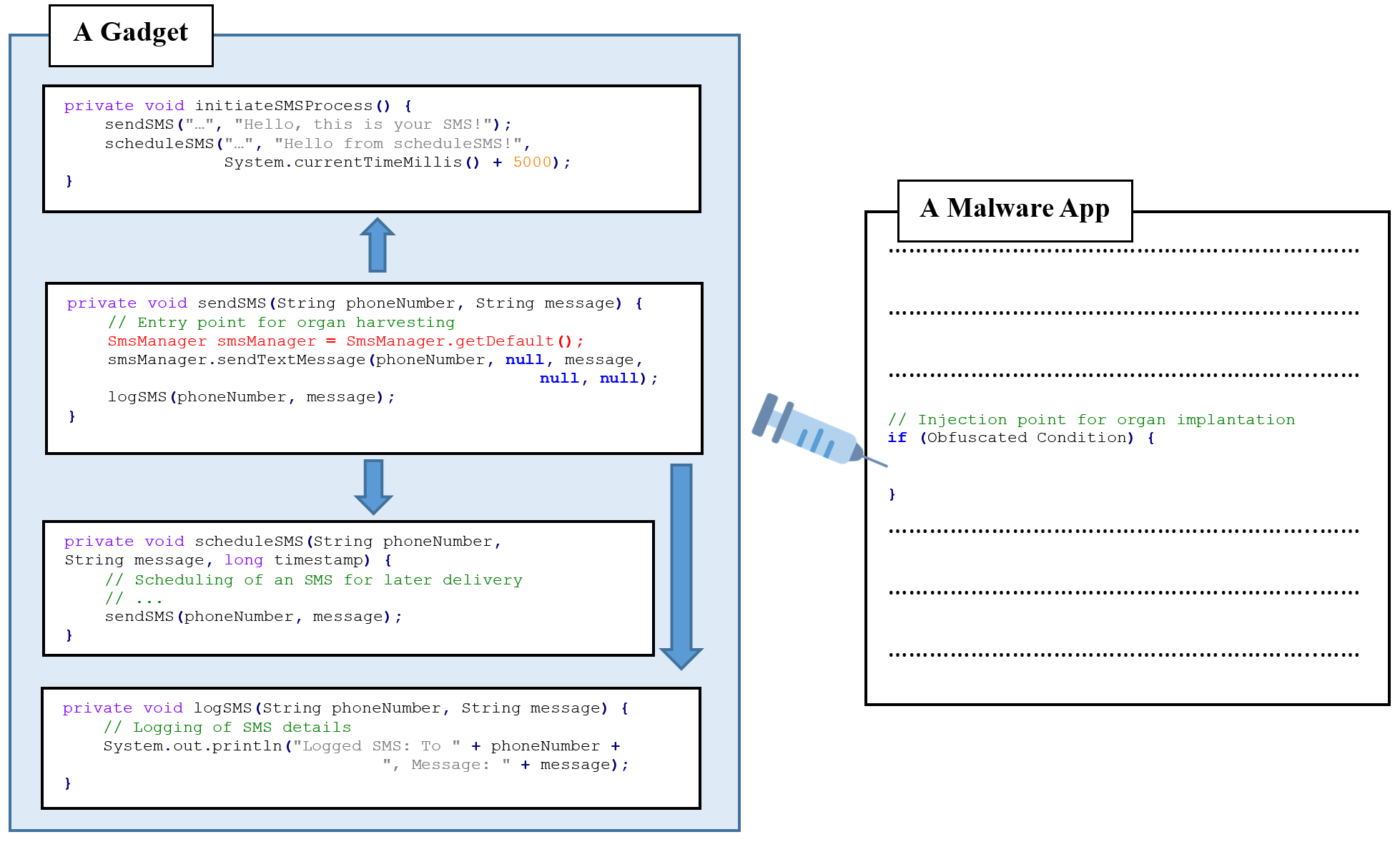}
    \caption{\ha{Applying a problem-space transformation (i.e., gadget) into a malware app involves injecting the gadget extracted from an API call entry point (\texttt{e.g., SmsManager}) in a donor into an obfuscated false condition statement within the malware app. The code snippets are displayed in \texttt{Java} representation to facilitate better understanding.}}
    \label{fig:gadget_injection}
\end{figure}

On the other hand, in the non-optimal hard-label attack, EvadeDroid applies random transformations to malware until it creates an AEs or reaches the predefined query budget. Specifically, in this setting, EvadeDroid randomly selects and applies a transformation from the action set to the malware app in each query. The target classifier is then queried to determine the label of the modified app. If the label indicates that the app is still classified as malware, EvadeDroid repeats this process.

\ha{It is important to highlight that Figure~\ref{fig:gadget_injection} depicts the procedure of manipulating an Android malware app through a problem-space transformation, specifically injecting an extracted gadget into a malware app.}
For more detailed information on the implementation of EvadeDroid, we refer the reader to~\ref{appendix:implementation_details}.
\section{Simulation Results}
\label{sec:simulation_results}
In this section, we empirically assess the performance of EvadeDroid in deceiving various academic and commercial malware classifiers. Our experiments aim to answer the following research questions:

\noindent\textbf{\hypertarget{RQ1}{RQ1}.} How does the evasion cost affect the performance of EvadeDroid? (\S\ref{sec:evasion_costs})

\noindent\textbf{\hypertarget{RQ2}{RQ2}.} Is EvadeDroid a versatile attack that can evade different Android malware detectors without relying on any specific assumptions? (\S\ref{sec:evasion_costs})

\noindent\textbf{\hypertarget{RQ3}{RQ3}.} How does the performance of EvadeDroid compare to other similar attacks? (\S\ref{sec:evadedroid_vs_other_attacks})

\noindent\textbf{\hypertarget{RQ4}{RQ4}.} Is EvadeDroid applicable in real-world scenarios? (\S\ref{sec:real_World_scenarios})

\noindent\textbf{\hypertarget{RQ5}{RQ5}.} How does EvadeDroid demonstrate its performance despite the restriction of not being able to query the target detectors? (\S\ref{sec:transferable_adversarial_examples})

\noindent\ha{\textbf{\hypertarget{RQ6}{RQ6}.} How does the proposed RS-based manipulation strategy affect the performance of EvadeDroid? (\S\ref{sec:search_strategy_impact})}

All experiments have been run on a Debian Linux workstation with an Intel (R) Core (TM) i7-4770K, CPU 3.50 GHz, and 32 GB RAM.

\subsection{Experimental Setup}
\label{subsection:experimental_setup}
Here, we provide an overview of the target detectors, datasets, and evaluation metrics we consider in our experiments.
\subsubsection{Target Detectors}
To ensure that our conclusions are not limited to a specific type of malware detection, we evaluate EvadeDroid against various malware detectors to demonstrate the effectiveness of the proposed attack. In particular, our evaluation focuses on assessing EvadeDroid's performance against well-known Android malware detection models, namely DREBIN~\cite{b30}, Sec-SVM~\cite{b19}, ADE-MA~\cite{b90}, MaMaDroid~\cite{b34}, and Opcode-SVM~\cite{b55}. These models have been extensively studied in the context of detecting problem-space adversarial attacks in the Android domain~\cite{b20,b26,b27,b75,zhang2021shadowdroid}. For more details about these detectors, please refer to~\ref{appendix:android_malware_detectors}.

\begin{table}[htbp]
\footnotesize
\begin{center}
\caption{Datasets used in our experiments.}
\begin{tabular}{cccc}
\toprule
{\textbf{Dataset}}&{\textbf{\shortstack{No. of \\Benign samples}}}&{\textbf{\shortstack{No. of \\Malware Samples}}}&{\textbf{\shortstack{Relevant \\Experiment}}}\\
\midrule
 \multirow{2}{*}{\textbf{\shortstack{Inaccessible Dataset \\(Training Samples)}}}
 & 10K & 2K & \S\ref{sec:evasion_costs}, \S\ref{sec:evadedroid_vs_other_attacks} ,\S\ref{sec:transferable_adversarial_examples}\\
 & 90K & 10K & \S\ref{sec:transferable_adversarial_examples}\\
\midrule
\textbf{\shortstack{Accessible Dataset \\(EvadeDroid's samples)}} & 2K & 1K & All\\
\bottomrule
\end{tabular}
\label{table:dataset}
\end{center}
\end{table}
\subsubsection{Dataset} 
We evaluate the performance of EvadeDroid using the dataset provided in~\cite{b26}. This dataset consists of $\approx 170K$ samples, each represented using the DREBIN~\cite{b30} feature set. The samples are feature representations of Android apps collected from AndroZoo~\cite{b73} and labeled by~\cite{b26} using a threshold-based labeling approach. These collected apps were published between January 2017 and December 2018. According to the labeling criteria in~\cite{b26}, an APK is considered malicious or clean if it has been detected by any $4+$ or 0 VirusTotal (VT)~\cite{b94} engines, respectively. It is important to note that the threshold-based labeling approach does not rely on specific engines but considers the number of engines involved~\cite{b92}. Therefore, the engines used for labeling may vary from sample to sample.

\begin{figure}[t]
    \centering
    \includegraphics[width=7.5cm]{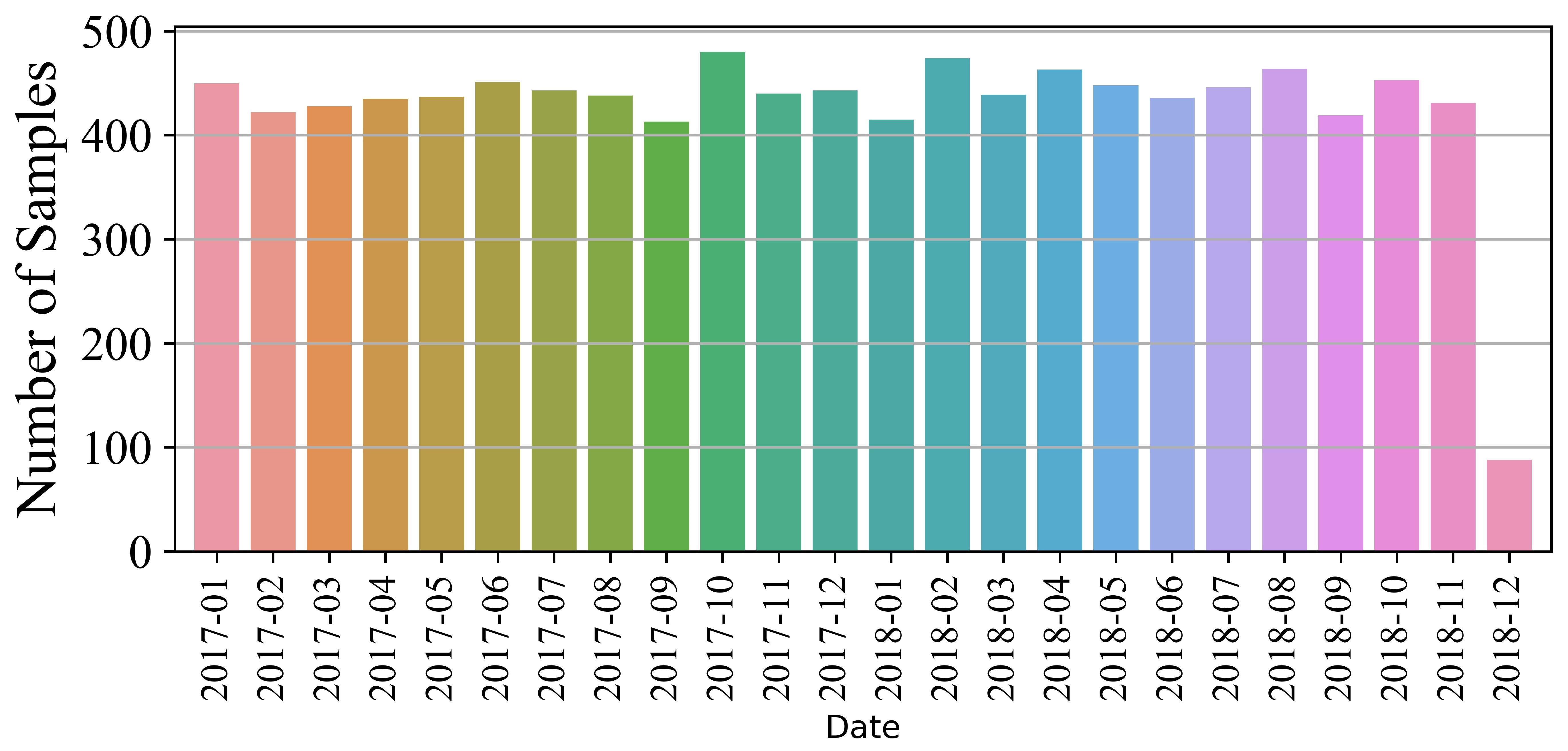}
    \caption{The temporal distribution of training samples. The dataset~\cite{b26} lacked clarity regarding the release dates of the $\approx 1.5K$ samples in our training set.}
    \label{fig:distribution_training_samples}
\end{figure}

Table~\ref{table:dataset} presents the specifications of datasets utilized in our research where their samples were randomly chosen from the collected data provided in~\cite{b26}. It's worth mentioning that there is no overlap between the inaccessible and accessible datasets. EvadeDroid exclusively makes use of the accessible dataset, which comprises $2K$ benign samples for donor selection and $1K$ malware samples for the creation of AEs. To fulfill the requirement of direct utilization of apps in our problem-space attack, we collect $3K$ apps corresponding to EvadeDroid's accessible samples from AndroZoo, based on the apps' specifications provided with the dataset~\cite{b26}. In this study, we employ two training sets with different scales (i.e., $12K$ and $100K$) for training classifiers. The proportion between benign and malware samples in the training sets is chosen to avoid spatial dataset bias~\cite{b74}. Figure~\ref{fig:distribution_training_samples} illustrates the temporal distribution of the smaller training set, demonstrating the absence of temporal bias as these apps were published across various months. The larger training set follows a similar distribution. In \S\ref{sec:evasion_costs}, \S\ref{sec:evadedroid_vs_other_attacks}, and \S\ref{sec:transferable_adversarial_examples}, a training set with a reasonable size (i.e., $12K$) is used due to the time-consuming preprocessing required by the apps in the MaMaDroid and Opcode-SVM, especially the former. Note that MaMaDroid and Opcode-SVM employ their own distinct feature representations, which differ from the DREBIN feature representation used in~\cite{b26}. Therefore, to provide the training set for these detectors, we have to directly collect all considered apps in the training set from AndroZoo based on the specifications provided by~\cite{b26}. Subsequently, the apps are embedded in the MaMaDroid and Opcode-SVM feature spaces using a feature extraction method. In the second evaluation conducted in \S\ref{sec:transferable_adversarial_examples}, we employ a larger training set (incl., $100K$ samples) to train DREBIN and Sec-SVM in order to illustrate the impact of a larger training set on EvadeDroid. It is important to highlight that our empirical evaluation shows that training classifiers with more samples does not significantly alter the performance of EvadeDroid.

\subsubsection{Evaluation Metrics} 
\begin{figure}[b]
    \centering
    \includegraphics[width=8cm]{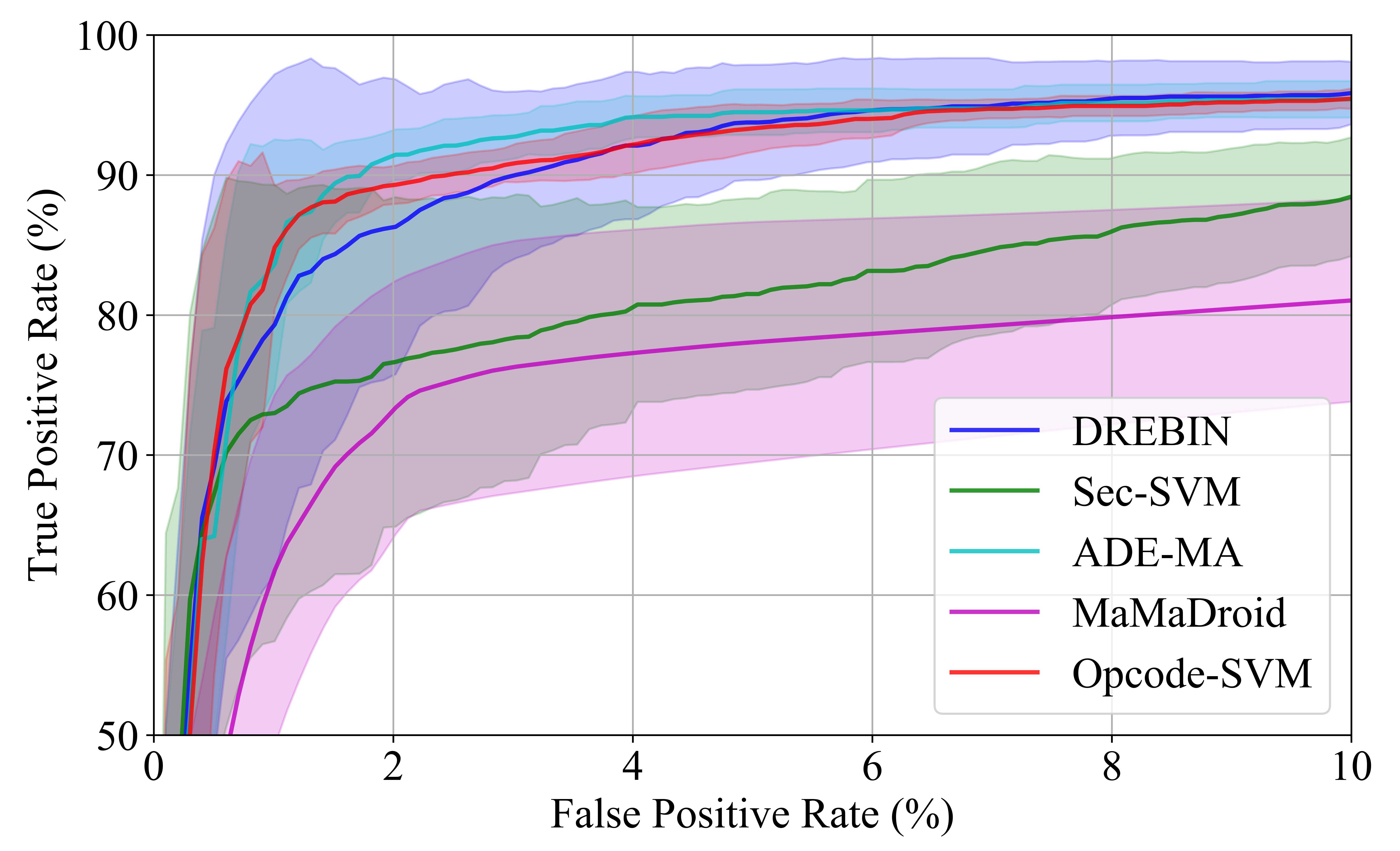}
    \caption{ROC curves of DREBIN, Sec-SVM, ADE-MA, MaMaDroid, and Opcode-SVM in the absence of adversarial attacks. The regions with translucent colors that encompass the lines are standard deviations.}
    \label{fig:ROC}
\end{figure}

We utilize the True Positive Rate (TPR) and False Positive Rate (FPR) as performance metrics for evaluating the effectiveness of malware classifiers in detecting Android malware. 
In Figure~\ref{fig:ROC}, we present the Receiver Operating Characteristic (ROC) curves of DREBIN, Sec-SVM, ADE-MA, MaMaDroid, and Opcode-SVM, the Android malware detectors used in this study, on the $12K$ training samples in the absence of our proposed attack. Note that the ROC curves were generated using 10-fold cross-validation. In addition to these metrics, we introduce the Evasion Rate (ER) \ha{and Evasion Time (ET)} as EvadeDroid's performance \ha{assessment metrics} in deceiving malware classifiers. ER is calculated as the ratio of correctly detected malware samples that are able to evade the target classifiers after manipulation to the total number of correctly classified malware samples. \ha{ET represents the average time,  expressed in seconds, required by EvadeDroid to generate an AE, encompassing both optimization and query times. Note that the optimization time primarily consists of the execution times of random search, injecting problem-space transformations, and performing feature extraction to represent manipulated apps within the feature space.} Further details of our experimental settings can be found in~\ref{appendix:experimental_settings}.

\begin{figure}[b!]
    \centering
    \includegraphics[width=\columnwidth]{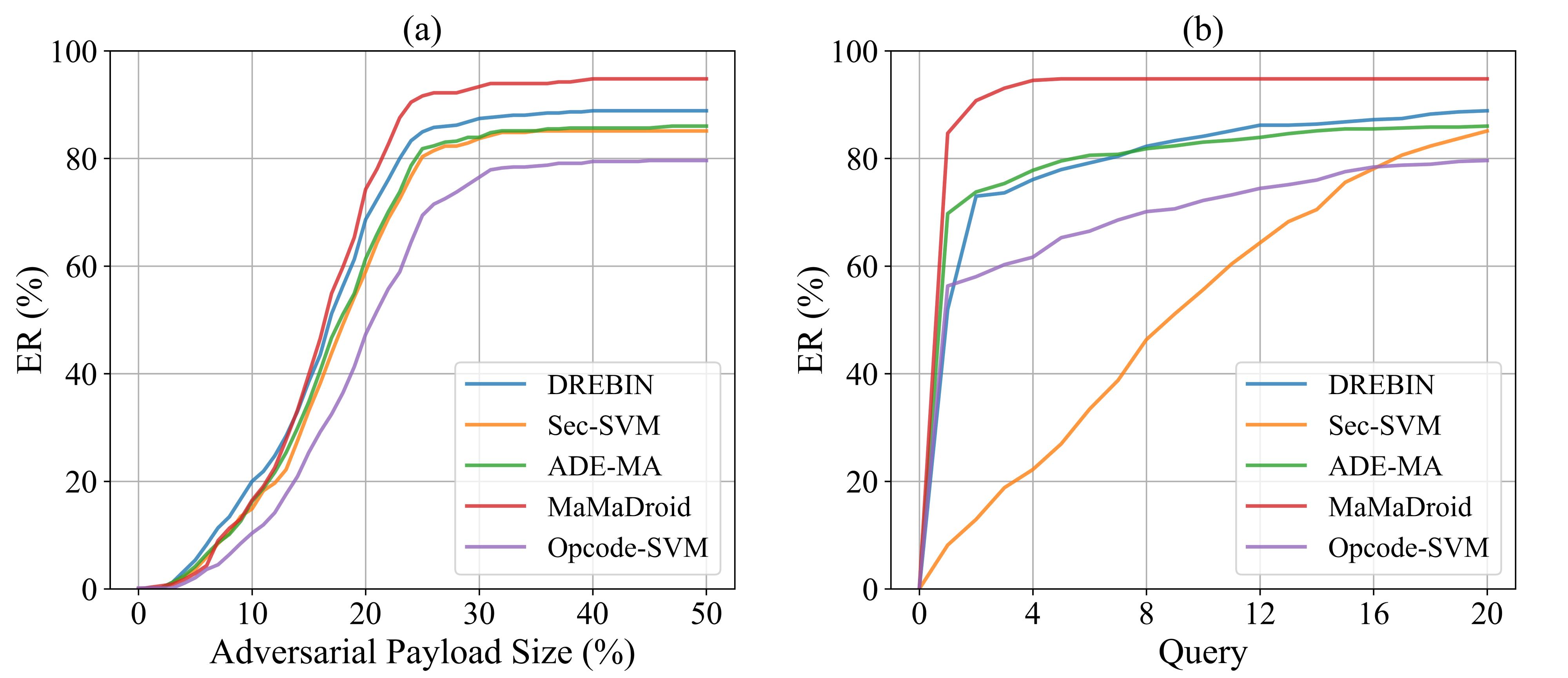}
    \caption{ERs of EvadeDroid operating in the soft-label setting in deceiving different Android malware detectors in terms of (a) different queries and (b) different adversarial payload sizes.}
    \label{fig:diff_evasion_costs}
\end{figure}

\subsection{Evasion Costs and Generalizability}
\label{sec:evasion_costs}
This section first examines the influence of the allowed adversarial payload size $\alpha$ and the query budget $Q$ on the performance of EvadeDroid to answer \textbf{\hyperlink{RQ1}{\textcolor{black}{RQ1}}}. Specifically, the evasion rates of EvadeDroid in fooling various malware detectors under different adversarial payload sizes and query numbers are depicted in Figure~\ref{fig:diff_evasion_costs}. Fig.~\ref{fig:diff_evasion_costs}a demonstrates that the evasion rate is influenced by the size of the adversarial payload, as increasing the size allows EvadeDroid to modify more malware applications. However, we observed that for $\alpha \geq 30\%$, the impact on the evasion rate becomes less significant, as most sequences of viable transformations almost reach a plateau at $\alpha = 30\%$. Furthermore, no further improvement in evasion rates is observed beyond $\alpha = 50\%$. In addition to the adversarial payload size, the query budget is another constraint that affects the evasion rate of EvadeDroid. Fig.~\ref{fig:diff_evasion_costs}b presents a comparison of the effect of different query numbers on the evasion rates of EvadeDroid against various malware detectors, with an allowed adversarial payload size of $\alpha = 50\%$. As can be seen in Fig.~\ref{fig:diff_evasion_costs}b, EvadeDroid requires a larger number of queries to generate successful AEs for bypassing Sec-SVM as compared to other detectors. This can be attributed to the fact that Sec-SVM, being a sparse classification model, relies on a greater number of features for malware classification compared to other classifiers. Consequently, EvadeDroid needs to apply more transformations to malware apps in order to deceive this more resilient variant of DREBIN. Additionally, Fig.~\ref{fig:diff_evasion_costs}b demonstrates that a query budget of $Q=20$ is nearly sufficient for EvadeDroid to achieve maximum evasion rate when attempting to bypass a malware detector. It is important to highlight that for the remaining experiments of the paper, we have chosen to use $Q=20$ and $\alpha = 50\%$ as they yield the optimal performance for EvadeDroid.

\begin{table}[t]
\footnotesize
\begin{center}
\caption{Effectiveness of EvadeDroid in misleading different malware detectors when $Q$ = 20 and $\alpha = 50\%$. NoQ, NoT, and AS denote Avg. No. of Queries, Avg. No. of Transformations, and Avg. Adversarial Payload Size, respectively.}
\label{table:cmp_evadedroid}
\begin{tabular}{ccccccc}
  \toprule
  \textbf{Type of Threat} & \textbf{Target Model} & \textbf{ER (\%)} & \ha{\textbf{ET (s)}} & {\textbf{\shortstack{NoQ}}}&{\textbf{\shortstack{NoT}}}&{\textbf{\shortstack{AS (\%)}}}\\ \hline
  \multirow{4}{*}{\textbf{Soft Label}}
   & DREBIN & 88.9 & \ha{210.3} & 3 & 2 & 15.5 \\ 
      & Sec-SVM & 85.1 & \ha{495.4} & 9 & 4 & 16.4\\ 
      & ADE-MA & 86.0 & \ha{126.2} & 2 & 1 & 16.3\\ 
      & MaMaDroid & 94.8 & \ha{131.4} & 1 & 1 & 15.9\\ 
      & Opcode-SVM & 79.6 & \ha{114.1} & 3 & 2 & 18.3\\ 
   \midrule
   \multirow{4}{*}{\textbf{\shortstack{Optimal\\ Hard Label}}}
      & DREBIN & 84.5 & \ha{240.6} & 4 & 2 & 16.2 \\
      & Sec-SVM & 82.6 & \ha{613.1} & 9 & 6 & 16.5 \\ 
      & ADE-MA & 84.4 & \ha{121.2} & 2 & 1 & 16.3 \\ 
       & MaMaDroid & 94.8 & \ha{133.7} & 1 & 1 & 15.9\\    
      & Opcode-SVM & 74.1 & \ha{101.2} & 2 & 1 & 18.2\\ 
    \midrule
    \multirow{4}{*}{\textbf{\shortstack{Non-optimal\\ Hard Label}}} 
      & DREBIN & 79.7 & \ha{357.2} & 4 & 4 & 16.9 \\ 
      & Sec-SVM & 78.2 & \ha{782.8} & 9 & 9 & 17.3 \\ 
      & ADE-MA & 82.7 & \ha{157.3} & 2 & 2 & 16.4 \\ 
      & MaMaDroid & 94.8 & \ha{132.6} & 1 & 1 & 15.9\\    
      & Opcode-SVM & 66.6 & \ha{76.2} & 1 & 1 & 18.3\\ 
      \bottomrule
    
\end{tabular}
\end{center}
\end{table}

To answer \textbf{\hyperlink{RQ2}{\textcolor{black}{RQ2}}}, we conduct an experiment involving various malware detectors and different attack settings. Specifically, we include DREBIN, SecSVM, ADE-MA, MaMaDroid, and Opcode-SVM to cover different ML algorithms (i.e., linear vs. non-linear malware classifiers, and gradient-based vs. non-gradient-based malware classifiers) and diverse features (i.e., discrete vs. continuous features, and syntax vs. opcode vs. semantic features). Additionally, we explore different attack settings (soft label vs. hard label) to demonstrate EvadeDroid's adaptability in various scenarios. The performance of the proposed attacks under different settings and malware detectors is presented in Table~\ref{table:cmp_evadedroid}. As shown in this table, EvadeDroid demonstrates effective evasion capabilities against various malware detectors, including DREBIN, Sec-SVM, and ADE-MA with syntax binary features, as well as MaMaDroid with semantic continuous features and Opcode-SVM with opcode binary features. The evaluation also reveals that EvadeDroid performs similarly well in the optimal hard-label setting compared to the soft-label setting. It is important to note that the comparison between soft-label attacking and non-optimal hard-label attacking highlights the influence of optimizing manipulations on the performance of EvadeDroid against different detectors. While only applying transformations to malware apps is sufficient for MaMaDroid, optimizing manipulations can enhance EvadeDroid's effectiveness against other detectors, especially Opcode-SVM. For instance, our findings shown in Table~\ref{table:cmp_evadedroid} demonstrate a $13\%$ improvement in the ER of EvadeDroid when targeting Opcode-SVM in the soft-label setting, compared to the non-optimal hard-label setting. Furthermore, when operating in the soft-label setting, EvadeDroid requires notably fewer transformations to bypass DREBIN and Sec-SVM, as compared to the non-optimal hard-label setting (e.g., 4 vs. 9 for Sec-SVM), which confirms the effectiveness of EvadeDroid in solving the optimization problem defined in equation~(\ref{eq_optimization1}). \ha{Table~\ref{table:cmp_evadedroid} further illustrates that our optimization leads to a substantial reduction in ET compared to the non-optimal hard-label setting. Specifically, for DREBIN and Sec-SVM, this leads to a time reduction of $\approx 41\%$ and $\approx 37\%$, respectively. This significant enhancement can be attributed to the reduction in the number of transformations, achieved through the utilization of our proposed optimization technique. Note that ET brings attention to the varying time overheads associated with the feature extraction process used to compute objective values when attacking different target detectors. For example, while NoQ and NoT are the same in attacking MaMaDroid and Opcode-SVM in the non-optimal hard-label setting, the ET for MaMaDroid is significantly higher than that for Opcode-SVM. The observed distinction is rooted in the considerable time consumption of the feature extraction process in MaMaDroid.}

In summary, the results demonstrate that the proposed adversarial attack is a versatile black-box attack that does not make assumptions about target detectors, including the ML algorithms or the features used for malware detection. Furthermore, it can operate effectively in various attack settings.

\subsection{EvadeDroid vs. Other Attacks}
\label{sec:evadedroid_vs_other_attacks}
To answer \textbf{\hyperlink{RQ3}{\textcolor{black}{RQ3}}}, we conduct an empirical analysis to assess how EvadeDroid performs in comparison to other similar attacks. 
To establish a comprehensive evaluation of EvadeDroid, we consider \ha{four} baseline attacks: PiAttack~\cite{b26}, Sparse-RS~\cite{b62}, ShadowDroid~\cite{zhang2021shadowdroid}, \ha{and GenDroid~\cite{xu2023gendroid}} operating in white-box, gray-box, semi-black-box, \ha{and black-box} settings, respectively. These attacks serve as suitable benchmarks, allowing us to assess the performance of EvadeDroid from different perspectives, such as evasion rate and the number of queries. \ha{Similar to} EvadeDroid, Sparse-RS, ShadowDroid, \ha{and GenDroid} generate AEs by querying the target detectors. Additionally, PiAttack is a problem-space adversarial attack that employs a similar type of transformation to generate AEs. \ha{Although PiAttack is a white-box evasion attack, it establishes a benchmark for optimal evasion performance, facilitating the evaluation of the comparative effectiveness of other attacks with limited or zero knowledge about the targeted detectors.} For further information about these attacks, please refer to~\ref{appendix:baseline_attacks}. In this experiment, we chose DREBIN, Sec-SVM, and ADE-MA as the target detectors because they align with the threat models of PiAttack, Sparse-RS, and ShadowDroid. Table~\ref{table:evadedroid_pk_random} shows the ERs of different adversarial attacks in deceiving various malware detectors. As can be seen in Table~\ref{table:evadedroid_pk_random}, although EvadeDroid has zero knowledge about DREBIN, Sec-SVM, and ADE-MA, its evasion rates for bypassing these detectors are comparable to PiAttack, where the adversary has full knowledge of the target detectors. Moreover, our empirical analysis shows that EvadeDroid requires adding more features to evade DREBIN, Sec-SVM, and ADE-MA. In concrete, on average, EvadeDroid makes 54--90 new features appear in the feature representations of the malware apps when it applies transformations to the apps for evading DREBIN, Sec-SVM, and ADE-MA, while the transformations used by PiAttack on average, trigger 11--68 features. PiAttack's ability to add a smaller number of features is attributed to its complete knowledge of the details of DREBIN, Sec-SVM, and ADE-MA. However, EvadeDroid lacks this specific information. 

\ha{Furthermore, as shown in Table~\ref{table:evadedroid_pk_random},} the evasion rate of Sparse-RS for DREBIN and Sec-SVM demonstrates that random alterations in malware features do not necessarily result in the successful generation of AEs, even when adversaries have access to the target models' training set. Although EvadeDroid operates solely in a black-box setting, this attack outperforms Sparse-RS by a considerable margin for both DREBIN and Sec-SVM, i.e., $70.6\%$ and $84.7\%$ improvement, respectively. \ha{Moreover, EvadeDroid considerably surpasses ShadowDroid in attacking Sec-SVM and ADE-ME. Especially,} in contrast to EvadeDroid, ShadowDroid is unsuccessful in effectively evading Sec-SVM, which is a robust detector against AEs. \ha{Note} that the superior performance of ShadowDroid compared to EvadeDroid in bypassing DREBIN is based on the assumption that target detectors primarily rely on API calls and permissions. However, this assumption is not practical in real scenarios, as detectors may employ other features for malware detection. \ha{Table~\ref{table:evadedroid_pk_random} further illustrates that GenDroid exhibits superior evasion rates compared to EvadeDroid when targeting DREBIN and ADE-MA; nevertheless, its efficacy is substantially nullified when facing Sec-SVM, a resilient malware detector.}
Our empirical analysis also highlights the remarkable efficiency of EvadeDroid in terms of the number of queries compared to other query-based attacks. Specifically, on average, EvadeDroid requires only 2--9 queries to bypass DREBIN, Sec-SVM, and ADE-MA, while Sparse-RS, ShadowDroid, \ha{and GenDroid} demand 2--195, 29--64, \ha{and 81--336} queries, respectively.

\begin{table}[t]
\footnotesize
\begin{center}
\caption{ERs of EvadeDroid, PiAttack, Sparse-RS, ShadowDroid, \ha{and GenDroid} in misleading DREBIN, Sec-SVM, and ADE-MA. \ha{NoQ denotes Avg. No. of Queries.}}
\label{table:evadedroid_pk_random}
\begin{tabular}{p{3cm}|p{3cm}cc}
  \toprule
  \textbf{Target Model} & \textbf{Evasion Attach} & \textbf{ER (\%)} & \textbf{NoQ} \\ \midrule
  \multirow{5}{*}{DREBIN} 
      & EvadeDroid & 88.9 & \ha{3}\\ 
      & PiAttack & 99.6 & \ha{N/A}\\
      & Sparse-RS & 18.3 & \ha{195}\\ 
      & ShadowDroid & 95.3 & \ha{31}\\
      & \ha{GenDroid} & \ha{95.5} & \ha{93}\\
       \midrule
   \multirow{5}{*}{Sec-SVM} 
      & EvadeDroid & 85.1 & \ha{9}\\ 
      & PiAttack & 94.3 & \ha{N/A}\\
      & Sparse-RS & 0.4 & \ha{38}\\ 
      & ShadowDroid & 8.6 & \ha{64}\\
      & \ha{GenDroid} & \ha{14.5} & \ha{336}\\
       \midrule   
     \multirow{5}{*}{ADE-MA} 
      & EvadeDroid & 86.0 & \ha{2}\\ 
      & PiAttack & 100 & \ha{N/A}\\
      & Sparse-RS & 99.7 & \ha{2}\\ 
      & ShadowDroid & 77.8 & \ha{29}\\
      & \ha{GenDroid} & \ha{100} & \ha{81}\\     
      \bottomrule
\end{tabular}
\end{center}
\end{table}




In summary, the experimental results validate the practicality of EvadeDroid, which adopts a realistic threat model, in comparison to other attacks for generating AEs. Specifically, the threat models of PiAttack and Sparse-RS are essentially proposed for the detectors that operate in the DREBIN feature space, but their threat models are not practical for targeting detectors like MaMaDroid. Furthermore, ShadowDroid's effectiveness is limited to scenarios where malware detection is solely based on API calls and permissions. For instance, as demonstrated in~\cite{zhang2021shadowdroid}, ShadowDroid is unable to deceive MaMaDroid or opcode-based detectors. In contrast, as shown in~$\S\ref{sec:evasion_costs}$, EvadeDroid is capable of effectively fooling these types of detectors as its problem-space transformations are independent of feature space. Additionally, \ha{although GenDroid operates in the ZK setting, it perhaps encounters challenges in evading robust malware detectors like Sec-SVM and might pose potential issues in real-world scenarios due to the substantial number of queries it requires compared to EvadeDroid. Finally,} Sparse-RS, ShadowDroid, \ha{and GenDroid might not be deemed realistic approaches} as their abilities to satisfy problem-space constraints, particularly robustness-to-preprocessing and plausibility constraints, are questionable.

\begin{table}[t!]
\footnotesize
\begin{center}
\caption{Performance of EvadeDroid in the hard-label setting on five commercial antivirus products. NoM denotes No. of Detected Malware by each engine among 100 malware apps.}
\label{table:cmp_evadedroid_on_vt}
\begin{tabular}{cccccc}
\toprule
\multirow{2}{*}{\textbf{Engine}} & \multirow{2}{*}{\textbf{\shortstack{NoM}}} & \multicolumn{4}{c}{\textbf{EvadeDroid}} \\
\cline{3-6}
 & & ER (\%) & \shortstack{Avg. Attack \\Time (s)} & \shortstack{Avg. No. of \\Queries} & \shortstack{Avg. Query\\ Time} \\
\midrule
 AV1 & 54 & 68.5 & 31.3 & 1 & 214.3\\

 AV2 & 32 & 87.5 & 54.7 & 2 & 387.2\\

 AV3 & 31 & 74.2 & 124.1 & 2 & 446.6\\

 AV4 & 41 & 100 & 35.2 & 1 & 329.7\\

 AV5 & 11 & 63.6 & 21.5 & 1 & 272.9\\
\bottomrule
\end{tabular}
\end{center}
\end{table}

\subsection{EvadeDroid in Real-World Scenarios}
\label{sec:real_World_scenarios}
\begin{figure}[t]
    \centering
    \includegraphics[width=8cm]{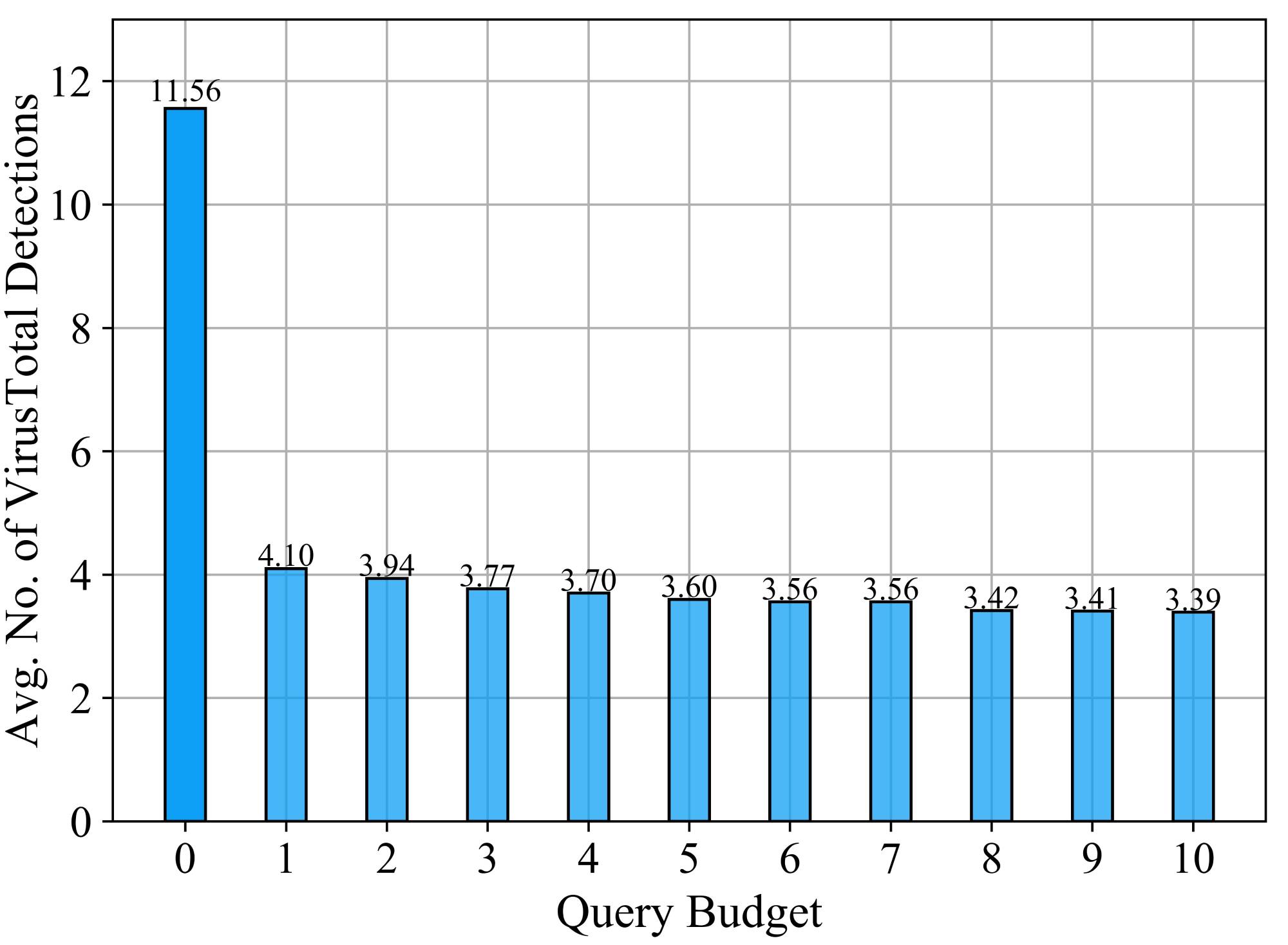}
    \caption{Performance of EvadeDroid in evading VT engines against different query budgets.}
    \label{fig:vt_overall_effect}
\end{figure}

This experiment aims to investigate \textbf{\hyperlink{RQ4}{\textcolor{black}{RQ4}}} to demonstrate the practicality of EvadeDroid in real-world scenarios. Although the ability of EvadeDroid in the hard-label setting indicates that this attack can transfer to real life, we further consolidate this observation by measuring the impact of EvadeDroid on commercial antivirus products that are available on VT to confirm the practicality of our proposed attack in real scenarios. We chose five popular antivirus engines in the Android ecosystem based on the recent ratings of the endpoint protection platforms reported by AV-Test~\cite{b95}. They are the top AVs in AV-Test capable of detecting malware apps in EvadeDroid's accessible dataset. Moreover, 100 malware apps belonging to different malware families have been randomly selected from the $1K$ malware apps available to EvadeDroid to evaluate the performance of this attack on the aforementioned five commercial detectors. To ensure the reliability of our experiment, it is crucial to confirm that the labels assigned to the malware apps used in this experiment have remained consistent. This is because the labels of collected apps are based on their corresponding samples in our benchmark dataset~\cite{b26}, while the labels assigned by antivirus engines to apps can potentially change over time. Therefore, we meticulously selected 100 apps that are still malware based on the threshold labeling criteria used in our primary dataset at the time of our experiment, i.e. on September 11, 2022, through querying VT. Furthermore, for each antivirus product, we generate AEs for the apps detected as malware by the antivirus.
Table \ref{table:cmp_evadedroid_on_vt} presents the results of the experiment in which EvadeDroid attempts to deceive each AV in the optimal hard-label setting. In this experiment, we have assumed $Q$~=~10 and $\alpha$~=~50\%. 
As can be seen in Table \ref{table:cmp_evadedroid_on_vt}, our proposed attack can effectively evade all antivirus products with a few queries. 
Here the effectiveness of EvadeDroid can be primarily attributed to the transformations rather than the optimization technique. This is evident from the fact that in most cases, only one query is required to generate AEs. We further investigate the performance of EvadeDroid against the overall effect of VT. 
Figure~\ref{fig:vt_overall_effect} shows the average number of VT detections for all $100$ malware apps after each attempt of EvadeDroid to change malware apps into AEs. As depicted in Fig.~\ref{fig:vt_overall_effect}, EvadeDroid can effectively deceive VT engines with an average of $70.67\%$. 
It is worth noting that the findings in this experiment validate the results observed in previous studies (e.g.,~\cite{ceschin2019shallow}). 

\noindent\textbf{Responsible Disclosure.} We conducted a responsible disclosure process to ensure the security community was informed of our research findings. As part of this process, we not only reached out to VT but also notified the antivirus engines that were affected by EvadeDroid by providing detailed information about our attack methodology and sharing some test cases.
\begin{table}[t]
\footnotesize
\begin{center}
\caption{Transferability of AEs generated by EvadeDroid.}
\label{table:transferability}
\begin{tabular}{p{3cm}|p{3cm}c}
  \toprule
  \textbf{Surrogate Model} & \textbf{Target Model} & \textbf{ER (\%)} \\ \midrule
  \multirow{4}{*}{DREBIN} 
      & Sec-SVM & 25.5\\ 
      & ADE-MA &  88.7\\
      & MaMaDroid & 63.0\\ 
      & Opcode-SVM &  42.2\\
       \midrule
   \multirow{4}{*}{Sec-SVM} 
      & DREBIN &   95.7\\
      & ADE-MA &  98.5\\
      & MaMaDroid & 95.4 \\ 
      & Opcode-SVM & 53.7\\
       \midrule   
     \multirow{4}{*}{ADE-MA} 
      & DREBIN &  49.3\\ 
      & Sec-SVM &  8.7\\ 
      & MaMaDroid &  67.5\\  
      & Opcode-SVM &  22.0\\
      \midrule
     \multirow{4}{*}{MaMaDroid} 
      & DREBIN &  41.1\\ 
      & Sec-SVM &  6.0\\ 
      & ADE-MA &  88.9\\ 
      & Opcode-SVM &  37.0\\
      \midrule
      \multirow{4}{*}{Opcode-SVM} 
      & DREBIN &  32.8\\ 
      & Sec-SVM &  10.9\\ 
      & ADE-MA &  66.8\\ 
      & MaMaDroid &  74.83\\
      \bottomrule
\end{tabular}
\end{center}
\end{table}

\subsection{Transferable Adversarial Examples}
\label{sec:transferable_adversarial_examples}
In general, when decision-based adversarial attacks, such as EvadeDroid, encounter difficulty in querying specific target detectors, they can create transferable AEs using a surrogate classifier. Here we explore \textbf{\hyperlink{RQ5}{\textcolor{black}{RQ5}}} by considering transferable AEs. To investigate the transferability of EvadeDroid, we evaluate the evasion rates of AEs generated on a model (e.g., Sec-SVM), which works as a surrogate model, in misleading other target models (e.g., DREBIN). This is a stricter threat model that indicates the performance of EvadeDroid in cases where adversaries are not capable of querying the target detectors. Table \ref{table:transferability} demonstrates that when EvadeDroid employs a stronger surrogate model (e.g., Sec-SVM), the AEs exhibit higher transferability. Note that the reported ERs in Table \ref{table:transferability} are the evasion rates of successful AEs that are also successfully transferred.

We further compare the transferability of EvadeDroid with PiAttack~\cite{b26} as it is similar to ours in terms of transformation type. 
This attack uses two kinds of primary features for misclassification, and side-effect features for satisfying problem-space constraints to generate realizable adversarial examples. However, EvadeDroid is not constrained by features as it operates in black-box settings. We specifically measure the transferability of the AEs in fooling Sec-SVM when DREBIN is the surrogate model. We ensure that the original apps of the AEs are correctly detected by Sec-SVM. Both DREBIN and Sec-SVM are trained with $100K$ apps (incl., $90K$ benign apps and $10K$ malware apps) to see the effect of large ML models on EvadeDroid's performance. The experimental results show that the ERs of the PiAttack and EvadeDroid in circumventing DREBIN are $99.06\%$ and $82.12\%$, respectively. Furthermore, EvadeDroid is much more transferable as the transferability of the AEs generated by EvadeDroid is $58.05\%$, while $23.23\%$ for PiAttack. 

\begin{table}[b]
\footnotesize
\begin{center}
\caption{\ha{RS-based vs. GA-based manipulation strategies in EvadeDroid. NoQ indicates Avg. No. of Queries and NoT denotes Avg. No. of Transformations.}}
\label{table:rs_vs_ga}
\begin{tabular}{p{3cm}|cccc}
  \toprule
  \textbf{\ha{Search Method}} &\textbf{\ha{ER (\%)}} & \textbf{\ha{ET (s)}} & \textbf{\ha{NoQ}} & \textbf{\ha{NoT}}\\ \midrule
  \ha{RS} & \ha{88.9} & \ha{210.3} & \ha{3} & \ha{2}\\   
  \ha{GA} & \ha{65.1} & \ha{630.7} & \ha{22} & \ha{5}\\     
  \bottomrule
\end{tabular}
\end{center}
\end{table}

\subsection{\ha{The impact of search strategy on EvadeDroid}}
\label{sec:search_strategy_impact}
\ha{To answer \textbf{\hyperlink{RQ6}{\textcolor{black}{RQ6}}}, we perform an empirical analysis to evaluate the performance of EvadeDroid when utilizing an alternative search strategy for manipulation. Specifically, we introduce a baseline manipulation method based on GA for use in EvadeDroid, where the fitness function of the baseline is the same as the RS-based method. In the proposed GA-based manipulation method, the individuals in the population (representing potential solutions) are binary strings with a length equal to the action set $\Delta$, where 1 indicates the corresponding transformation in $\Delta$ should be used for manipulation. This approach enhances the solution across various generations. In this experiment, the query budget for GA is set at 50 due to scalability concerns, as evaluating more solutions obtained by applying different sequences of transformations to the malware app would significantly increase time overheads. Moreover, our preliminary experiment suggests considering 9 as the population size of the GA-based method. Note that a large population size negatively affects the performance of GA, as the perturbation budget is quickly consumed by individuals in the initial generations.}

\ha{Table~\ref{table:rs_vs_ga} presents the results of the baseline when DREBIN is the target malware detector. As shown in Table~\ref{table:rs_vs_ga}, using RS in EvadeDroid outperforms GA. Specifically, RS not only leads to a 36.5\% enhancement in ER but also accelerates EvadeDroid by $\approx3\times$. These improvements are achieved with only 3 queries compared to GA's 22 queries.} 

\subsection{Discussion}
\label{sec:discussion}
\noindent\textbf{Real-world applicability.} 
EvadeDroid demonstrates its ability to generate practical adversarial Android apps by considering real-world attack limitations, such as operating in \ha{ZK} settings. We assume that EvadeDroid has no knowledge about the target malware classifiers and can only query them to obtain the labels of Android apps. Additionally, in some experiments, we assume that the target malware detectors only provide hard labels in response to the queries. The performance of EvadeDroid in various experiments validates its practicality. In a hard-label setting, it efficiently evades five popular commercial antivirus products with an average evasion rate of nearly 80\%. Furthermore, empirical evaluations of EvadeDroid on DREBIN, Sec-SVM, ADE-MA, MaMaDroid, and Opcode-SVM result in evasion rates of 89\%, 85\%, 86\%, 95\%, and 80\%, respectively. The success of our attack can be attributed to our approach of directly crafting adversarial apps in the problem space rather than perturbing features in the classifier's feature space. 
From a defender's perspective, EvadeDroid can be utilized in adversarial retraining to enhance the robustness of Android malware detection against realistic evasion attacks. ~\ref{data_augmentation} includes an experiment showcasing the adversarial robustness that can be achieved with the involvement of EvadeDroid.

\noindent\textbf{Functionality preserving.} We extended the tool presented in~\cite{b26}, in particular the organ-harvesting component, to manipulate malware apps. This tool ensures the preservation of functionality by adding dead codes to malware apps without affecting their semantics. Specifically, it incorporates opaque predicates, an obfuscated condition, to inject adversarial payloads into the apps while remaining unresolved during analysis, ensuring the payloads are never executed. Generally, verifying the semantic equivalence of two programs (e.g., a malware app and its adversarial version) is not trivial~\cite{b35}. Therefore, similar to the prior studies~\cite{b26,b90,b29}, our primary goal is to consider the installability and executability of apps to verify the correct functioning of the adversarial apps. To this end, we developed a scalable test framework that installs and executes adversarial apps on an Android Virtual Device (AVD) and conducts monkey testing~\cite{monkeytest} to simulate random user interactions with the apps to guarantee the stability of the apps. Furthermore, taking inspiration from prior research~\cite{li2023black}, we incorporate a log statement within the opaque predicate to ensure that the functionality of the manipulated apps remains unchanged. By monitoring the absence of log outputs, we can ascertain that the injected payloads are not executed. We select 50 adversarial apps, representing diverse malware families, for which their original malware apps can be installed and executed on the AVD without any issues. These apps are then subjected to our test framework. While the flaws in the Soot~\cite{b47} framework (e.g., the injection of payloads through Soot might result in incorrect updates to the function address table of the app), utilized in the manipulation tool~\cite{b26}, affect the executability of a few cases, the majority of the apps passed the test. 

\noindent\textbf{Query efficiency.} According to the experimental results obtained by applying EvadeDroid on academic and commercial malware detectors, we demonstrated that it can successfully carry out a query-efficient black-box attack. For instance, our proposed attack often only needs an average of 4 queries to generate the AEs that can successfully bypass DREBIN, Sec-SVM, ADE-MA, MaMaDroid, and Opcode-SVM. Moreover, we showed that EvadeDroid can effectively fool commercial antivirus products with less than two queries. One of the main reasons for being a query-efficient attack is due to the well-crafted transformations gathered in the action set. To maintain EvadeDroid's performance, it is crucial to periodically update the action set by incorporating newly published apps as new potential donors.
Besides the quality of the action set, the presented optimization method is another important aspect of our proposed attack that can facilitate the identification of an optimal sequence of transformations, especially when the target detectors are robust to AEs (e.g., Sec-SVM). In fact, the proposed RS technique is an efficient \ha{sampling-driven} search strategy that can quickly converge to a proper solution. \ha{Table~\ref{table:search_strategy} shows that Android evasion attacks often employ gradient-driven (e.g., gradient descent) and sampling-driven (e.g., GA) methodologies, where the latter is more practical for black-box evasion attacks because they can overcome the challenges inherent in using gradient-driven attacks in ZK settings. Specifically, gradient-driven attacks require access to precise details of target malware detectors and are limited to differentiable-based classifiers, which are not applicable to attacks operating in ZK settings. Moreover, gradient-driven techniques are not well-suited for continuous features, whereas the malware domain predominantly involves discrete features. Ultimately, gradient-masking~\cite{papernot2016practical} defenses implemented in target malware detectors demonstrate effectiveness in preventing gradient-driven attacks. It is important to note that our proposed sampling-driven method demonstrates greater efficiency compared to query-based methods used in other studies (e.g.,~\cite{he2023efficient,xu2023gendroid,b62,b23}). For instance, as shown in \S\ref{sec:evadedroid_vs_other_attacks}, EvadeDroid can effectively evade DREBIN with only 3 queries, whereas GenDroid needs 93 queries. Additionally, as illustrated in \S\ref{sec:search_strategy_impact}, our proposed RS-based strategy requires $\approx 210$ seconds to bypass DREBIN, while the GA-based methods extend the evasion time to $\approx 631$ seconds.}

\begin{table}[t!]
\footnotesize
\begin{center}
\caption{\ha{The prevalent search strategies employed in Android evasion attacks.}}
\label{table:search_strategy}
\begin{tabular}{p{3cm}p{6cm}c}
\toprule
{\textbf{\ha{Search Strategy}}} & {\textbf{\ha{Description}}} & {\textbf{\ha{Study}}} \\
\midrule
 \ha{Gradient-driven} & \ha{Utilizes gradients to iteratively adjust perturbations towards optimal adversarial perturbations.} & \ha{\cite{zhang2021shadowdroid,b19,b20,b21,b26,b27,b75,b90}}\\
 \midrule
 \ha{Sampling-driven} & \ha{Involves exploring the solution space by sampling candidate perturbations to find optimal adversarial perturbations.} &  \ha{\cite{xu2023gendroid,he2023efficient,li2023black,b62,b17,b22,b23,b29}}\\
\bottomrule
\end{tabular}
\end{center}
\end{table}
 \noindent \ha{\textbf{Scalability and effectiveness.} Our empirical evaluations demonstrate the ability of the EvadeDroid to adapt and work effectively across a large scale of targets. Especially the results in \S\ref{sec:evasion_costs}} highlight the effectiveness of our evasion attack in bypassing diverse malware \ha{detectors (i.e., linear vs. non-linear malware classifiers, and gradient-based vs. non-gradient-based malware classifiers)} that utilize different features (i.e., syntax, opcode, and semantic features) with different feature types (i.e., discrete and continuous features). \ha{
Furthermore, although manipulating applications within the problem space is inherently a time-consuming endeavor, the efficiency in querying allows our attack to autonomously generate AEs at a good speed, eliminating the need for manual and labor-intensive methods. Our empirical assessment in \S\ref{sec:transferable_adversarial_examples} also demonstrates that AEs generated by EvadeDroid to target a specific detector exhibit reusability across various malware detectors.}

\noindent \ha{\textbf{Potential applications.} EvadeDroid shows promise for various real-world applications within the realm of Android malware detection. Security professionals and organizations involved in the development and deployment of malware detectors can utilize EvadeDroid for security testing and evaluation. For instance, they can simulate adversarial scenarios to identify vulnerabilities and enhance the robustness of their systems against real-world threats. The adversarial training capabilities of the system render EvadeDroid a helpful asset for developers seeking to strengthen malware detectors against real-world evasion attacks. Moreover, our attack can be instrumental in the development of countermeasures, allowing cybersecurity experts to understand and address potential weaknesses in existing malware detection systems.}

\section{Limitations and Future Work}
\label{sec:limitations_and_future_work}
In this section, we elaborate on the limitations of our proposed method, which can be considered as future work. One of the concerns of EvadeDroid is the adversarial payload size (i.e., the relative increase in the size of AEs) that might be relatively high, especially for the small Android malware apps. This deficiency may cause malware detectors to be suspicious of the AEs, particularly for popular Android applications. Improving the organ harvesting used in the program slicing technique, in particular, finding the smallest vein for a specific organ, can address this limitation as each organ has usually multiple veins of different sizes. 


Additionally, EvadeDroid particularly crafts malware apps to mislead the malware detectors that use \textit{static} features for classification. We do not anticipate our proposed evasion attack to successfully deceive ML-based malware detectors that work with behavioral features specified by dynamic analysis as the perturbations are injected into malicious apps within an \texttt{IF} statement that is always \texttt{False}. Therefore, it remains an interesting avenue for future work to evaluate how our proposed attack can bypass behavior-based malware detectors.

\ha{Furthermore}, since EvadeDroid uses a well-defined optimization problem \ha{outlined in Algorithm~\ref{alg_generate_adm}}, it can be extended to other platforms \ha{(e.g., Windows)} if attackers \ha{offer problem-space transformations that are tailored to manipulate real-world objects (e.g., Windows Portable Executable files)}. This is because the transformations used in EvadeDroid can only be applied to manipulate Android applications. We leave further exploration as future work since it is beyond the scope of this study.

\ha{Finally, our study comprehensively covers various malware detection systems, employing diverse classifiers on different features with various types. However, there is an opportunity to improve the validity of our findings since the evaluation is conducted in controlled laboratory settings. Future research should delve deeper into the applicability of our adversarial attack framework in real-world environments, where dynamic factors like evolving malware landscapes and deployment scenarios may impact the attack's performance.}

\section{Conclusions}
\label{sec:conclusions}

This paper introduces EvadeDroid, a novel Android evasion attack in the problem space, designed to generate real-world adversarial Android malware capable of evading ML-based Android malware detectors in a black-box setting. Unlike previous approaches, EvadeDroid directly operates in the problem space without initially focusing on finding feature-space perturbations. Experimental results demonstrate the effectiveness of EvadeDroid in deceiving various academic and commercial malware detectors.


\section*{Acknowledgement}
Veelasha Moonsamy was funded by the Deutsche Forschungsgemeinschaft (DFG, German Research Foundation) under Germany’s Excellence Strategy - EXC 2092 CASA - 390781972.
\bibliographystyle{elsarticle-main} 
\bibliography{cas-refs}
\biboptions{sort&compress}
\appendix
\label{appendix}


\section{Problem-Space Constraints}
\label{appendix:problem_space_constraints}
To generate realizable AEs, adversarial attacks need to consider the following four problem-space constraints~\cite{b26}:
\begin{itemize}    
    \item \textbf{Available transformations} describe the types of manipulations (e.g., adding dead codes) that an adversary can utilize to modify malware apps.
    \item \textbf{Preserved semantics} constraint explains that the semantics of an Android app should be maintained after applying a transformation to the app.
    \item \textbf{Robustness-to-preprocessing} constraint describes the requirement that non-ML methods (e.g., preprocessing operators) should not be able to undo the adversarial changes.
    \item \textbf{Plausibility} constraint explains adversarial apps must look realistic (i.e., naturally created) under manual inspection.
\end{itemize}

\section{Donors Evaluation}
\label{appendix:donors_evaluation}
In this evaluation, we assess the influence of our donor selection strategy on the performance of EvadeDroid. Two action sets, denoted as $\Delta_1$ and $\Delta_2$, are provided, each containing 20 transformations. The transformations in $\Delta_1$ and $\Delta_2$ are chosen at random from the collection of transformations extracted from the 10 most similar apps and the 10 least similar apps to malware apps, respectively. To understand the process of finding similar apps, refer to Section \ref{section:preparation}. We then use these action sets in EvadeDroid to transform 50 randomly selected malware apps into AEs. Table~\ref{table:donors_evaluation} presents a comparison of the impact of $\Delta_1$ and $\Delta_2$ on EvadeDroid's performance. As can be seen in this table, when using $\Delta_1$, the number of queries and transformations is significantly reduced compared to $\Delta_2$. This finding validates that leveraging benign apps that resemble malware apps as the donors of transformations can reduce the cost of generating AEs, specially in terms of the required queries. 

\begin{table}[ht]
\fontsize{8}{10}\selectfont 
\begin{center}
\caption{The performance of EvadeDroid in attacking DREBIN when it utilizes two different action sets $\Delta_1$ and $\Delta_2$.}
\label{table:donors_evaluation}
\begin{tabular}{c|ccc}
\toprule
{\textbf{Action Set}}&{\textbf{ER~(\%)}}&{\textbf{\shortstack{Avg. No. of\\ Queries}}}&{\textbf{\shortstack{Avg. No. of \\Transformations}}}\\
\midrule
$\Delta_1$ & 66.0 & 3&2\\
$\Delta_2$ & 68.0 & 7&3\\
\bottomrule
\end{tabular}
\end{center}
\end{table}

\section{Implementation Details}
\label{appendix:implementation_details}
\begin{figure*}[t]
    \centering
    \includegraphics[width=11cm]{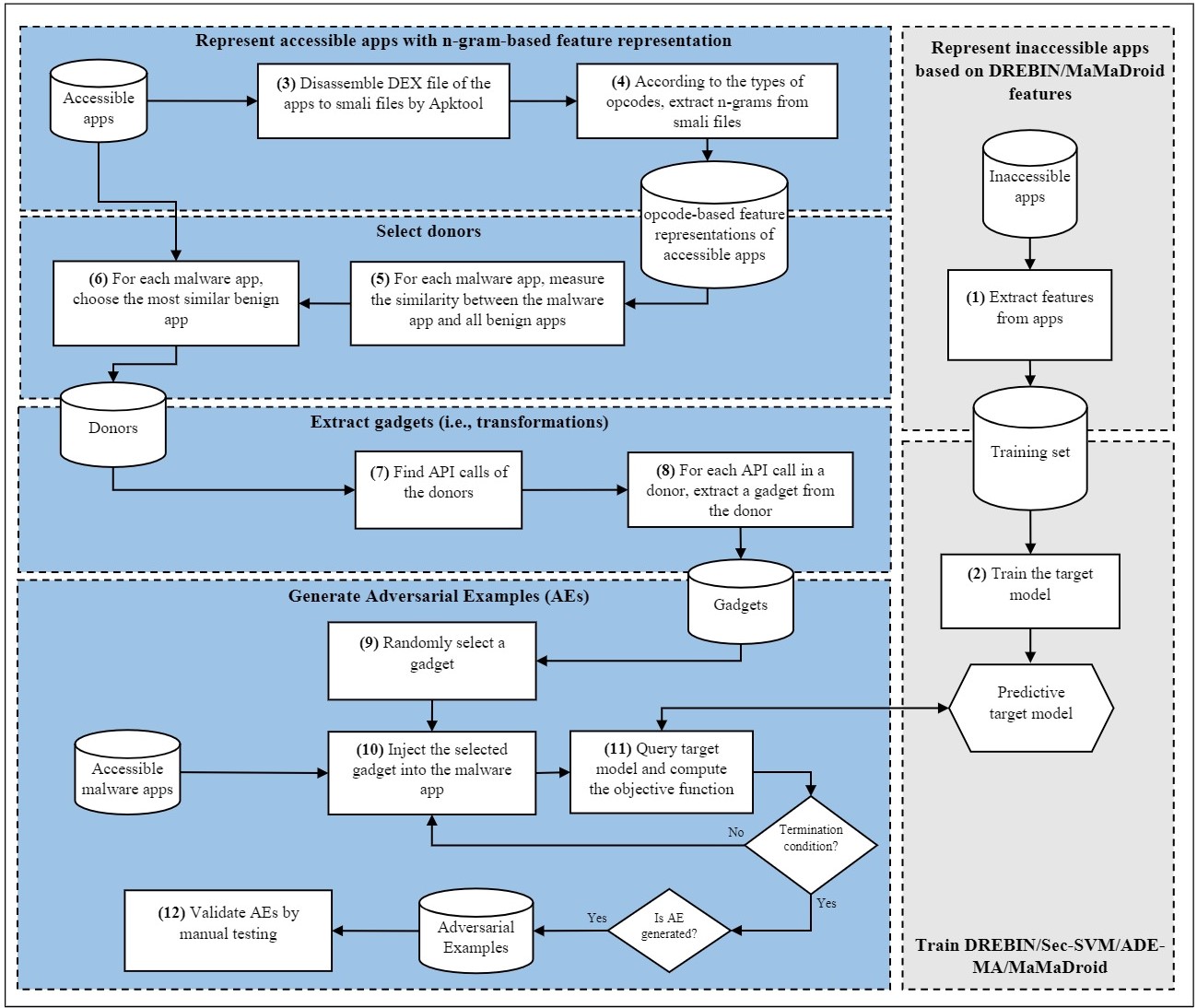}
    \caption{The details of the proposed framework. The blue and gray areas represent the workflows of EvadeDroid and target black-box malware detection, respectively.}
    \label{fig:evadedroid_pipeline}
\end{figure*}

The proposed framework illustrated in Figure~\ref{fig:evadedroid_pipeline} is implemented with Python 3 and Java 8. The source code\footnote{\href{https://github.com/HamidBostani2021/EvadeDroid}{https://anonymous.4open.science/r/EvadeDroidMain-1E69}} of the pipeline has been made publicly available to allow reproducibility. The components of EvadeDroid's pipeline are clearly depicted in Figure~\ref{fig:evadedroid_pipeline}. This section reviews some of the components that have not been previously described in detail in the paper.

\begin{itemize}
    \item \textbf{Component 7.} To identify API calls in donor apps, we utilize the tool provided in~\cite{b96}. This tool leverages Apktool~\cite{b46} to access the DEX files of Android apps, which are represented as smali files. It employs string analysis techniques to scan these files and identify the API calls present within them.
    \item \textbf{Component 8.} We extend the tool presented in~\cite{b26} to extract API calls from donors because this tool, which is based on the Soot framework, originally harvests Activities and URLs only.
    \item \textbf{Component 10.} The tool presented in~\cite{b26} has also been used to inject gadgets into malware apps (i.e., hosts). This tool ensures the fulfillment of both the preserved-semantic and robustness-to-preprocessing constraints by utilizing opaque predicates~\cite{b49} for transplanting the gadgets into hosts. The opaque predicates employed in the tool are obfuscated condition statements that encapsulate the injected gadgets. During runtime, these statements always evaluate to \texttt{False}, thereby preserving the semantics of malware apps as the injected gadgets remain unexecuted. Furthermore, the preprocessing operators are unable to eliminate the injected gadgets as the result of the statement cannot be statically resolved from the source code during design time. It is important to note that the generated AEs are plausible. \ha{This is because} the manipulation of malware apps involves the injection of realistic gadgets \ha{found} in benign apps. \ha{Additionally, the injection of the gadget occurs in unnoticeable injection points, maintaining the homogeneity complexity of the host's components.} The inclusion of gadgets may enhance EvadeDroid's performance by introducing more features in the manipulated apps. For further insights into the tool, we refer readers to~\cite{b26}.
\end{itemize}

\section{Android Malware Detectors}
\label{appendix:android_malware_detectors}
\noindent\textbf{DREBIN}~\cite{b30} and \textbf{Sec-SVM}~\cite{b19} are two prominent approaches in Android malware detection. DREBIN utilizes binary static features and employs linear Support Vector Machine (SVM) for classification. It extracts various features, including requested permissions and suspicious API calls, from the Manifest and DEX files of APKs through string analysis~\cite{b44}. These features are then used to construct a feature space for the classifier. In DREBIN, each app is represented by a sparse feature vector, where each entry indicates the presence or absence of a specific feature. Secure SVM (Sec-SVM) is an enhanced version of DREBIN that aims to enhance the resilience of linear SVM against adversarial examples. The core concept behind Sec-SVM is to increase the cost of evading the model when generating adversarial examples. Compared to DREBIN, Sec-SVM relies on a larger set of features for malware detection, making it more challenging to evade. Since Sec-SVM is a sparse classification model, it leverages a greater number of features to improve its malware detection capabilities

\noindent\textbf{ADE-MA}~\cite{b90} is an ensemble of deep neural networks (DNNs) that is strengthened against adversarial examples with \textit{adversarial training}. The adversarial training method tunes the DNN models by solving a min-max optimization problem, in which the inner maximizer generates adversarial perturbations based on a mixture of attacks, i.e. iterative “max” Projected Gradient Descent (PGD) attacks.

\noindent\textbf{MaMaDroid~\cite{b34}} utilizes static analysis to detect Android malware. The goal of MaMaDroid is to capture the semantics of an Android app by employing a Markov chain based on abstracted sequences of API calls. The process begins with generating a call graph for each Android app. From this call graph, the sequences of API calls are extracted and abstracted into different modes, including families, packages, and classes. Subsequently, MaMaDroid constructs a Markov chain for each abstracted API call in an APK, where each state represents a family, package, or class, and the transition probabilities indicate the state transitions. Finally, feature vectors incorporating continuous features are created based on the generated Markov chains.

\noindent\textbf{Opcode-SVM}~\cite{b55} is an Android malware detection method that utilizes static opcode-sequence features instead of predefined features. This approach focuses on performing n-gram opcode analysis to represent apps in a feature space, where a malware classifier is constructed. Specifically, the method employs a linear SVM with 5-gram binary opcode features to effectively detect Android malware.

\section{Experimental Settings}
\label{appendix:experimental_settings}
\noindent\textbf{Android malware detectors.} We built DREBIN, Sec-SVM, MaMaDroid, and ADE-MA based on their available source codes (i.e., ~\cite{b97,b98,b99}) that have been published in online repositories. Moreover, we have reproduced Opcode-SVM based on the implementation details provided in~\cite{b55}. The hyperparameters of the reproduced malware detectors are similar to those considered in their original studies~\cite{b26,b34,b90,b55}. Note that in our paper, the reproduced MaMaDroid~\cite{b34} is based on the K-Nearest Neighbors (KNN) algorithm with $k = 5$. This malware classifier operates in the family mode in all experiments. KNN algorithm is used in MaMaDroid as we empirically concluded that KNN performs better on our dataset than other classifiers employed in~\cite{b34}.

\noindent\textbf{Baseline evasion attacks.} We implemented Sparse-RS, ShadowDroid, \ha{and GenDroid} with Python~3 based on their relevant studies \ha{(i.e., ~\cite{b62,zhang2021shadowdroid,xu2023gendroid})}. Moreover, PiAttack~\cite{b26} has been built based on their available source codes published in~\cite{b97}.

\noindent\textbf{EvadeDroid.} Besides query budget $Q$ and the allowed adversarial payload size $\alpha$ that have been mentioned earlier, $n$ is another hyperparameter that shows the length of overlapping sub-string of opcodes' types in $n$-gram-based feature extraction. In this study, we consider $n=5$ because in ~\cite{b55}, the authors have shown that the best classification performance for opcode-based Android malware detection can be achieved with the 5-gram features. Furthermore, we select the top-$100$ benign apps as suitable donors for gadget extraction. Note that we consider $100$ donors as organ harvesting from donors is a time-consuming process.  


\section{Baseline Attacks}
\label{appendix:baseline_attacks}
\noindent\textbf{PiAttack}~\cite{b26} is a white-box attack in the problem space that generates real-world AEs using transformations called gadgets. This attack comprises two main phases: the initialization phase and the attack phase. In the initialization phase, key benign features are identified, and then gadgets corresponding to the identified features are collected from benign apps. In the attack phase, a greedy search strategy is used to find optimal perturbations by selecting gadgets based on their contribution to the feature vector of the malware app. This process is repeated until the modified feature vector is classified as a benign sample. Note that PiAttack incorporates both primary features and side-effect features into malware apps. The primary features are added to bypass detection, while the side-effect features are included to meet problem-space constraints.

\noindent\textbf{Sparse-RS}~\cite{b62} attack is a soft-label attack that gradually converts malware samples into AEs by querying the target model. Sparse-RS, which is a gray-box attack in the malware domain, finds the \emph{$l_{0}$}-bounded perturbations (i.e., the maximum allowed perturbations) via random search. Note that we set initial decay factor $\alpha_{init}=1.6$ and sparsity level $k=180$ similar to~\cite{b62} and query budget $Q=1000$.

\noindent\textbf{ShadowDroid}~\cite{zhang2021shadowdroid} is a black-box problem-space attack that generates AEs by building a substitute classifier, which is a linear SVM. The substitute classifier is built on binary feature space compromised by permissions and API calls. This attack makes a key feature list based on the importance of features specified by the substitute classifier. The attack adds the key features to a malware app and queries the target classifier to check if the manipulated app is classified as malware. ShadowDroid continues this process until reaching the maximum query budget or generating an AEs. We set query budget $Q=100$, following a similar setting as in \cite{zhang2021shadowdroid}. Note that ShadowDroid is not fully compatible with the zero-knowledge (ZK) setting as it relies on the assumption that the target detectors utilize permissions and API calls for malware detection. 
However, since it is a query-based problem-space attack, it serves as a proper naive problem-space baseline attack for our study.

\noindent\ha{\textbf{GenDroid~\cite{xu2023gendroid}} is a black-box Android evasion attack building upon GenAttack~\cite{alzantot2019genattack}. This query-based attack utilizes GA to discover adversarial perturbations in soft-label settings. GenDroid extends GenAttack by redesigning the fitness function, adopting a new evolutionary strategy, and incorporating Gaussian Process Regression (GPR) to guide evolution. Specifically, the fitness function is defined through a logarithmic transformation, incorporating adjustable weight parameters ($\alpha$ and $\beta$) and a norm-bounded perturbation. The selection process prioritizes elite individuals with higher fitness scores, and the \textit{Softmax} function is employed to convert fitness scores into probabilities. GPR is introduced to predict fitness values for individuals in the next generation. We empirically set the population size to 8 and the maximum number of generations to 50.}

\section{Data Augmentation}
\label{data_augmentation}
\begin{table}[b!]
\fontsize{8}{10}\selectfont 
\begin{center}
\caption{The impact of various training strategies on the utility of DREBIN.}
\label{table:performance_drebin_augmented_drebin}
\begin{tabular}{l|ccc}
\toprule
{\textbf{Model}}&{\textbf{No. of AEs}}&{\textbf{TPR~(\%)}}&{\textbf{FPR~(\%)}}\\
\midrule
Standard Training &N/A& 80.8 & 1.7\\
\midrule
\multirow{4}{*}{Adversarial Re-training} &500& 78.3 & 1.4\\
&1000& 74.9 & 0.9\\
&1500& 68.7 & 0.5\\
&1769& 32.7 & 0.2\\
\bottomrule
\end{tabular}
\end{center}
\end{table}
In this experiment, we evaluate the performance of EvadeDroid in enhancing the adversarial robustness of Android malware detection. To achieve this, we transform malware samples from the original training set into AEs using EvadeDroid. Subsequently, we re-train DREBIN using the modified dataset, resulting in a model that is robust to EvadeDroid. Our empirical analysis demonstrates that incorporating AEs generated by EvadeDroid in the training set of DREBIN can effectively thwart the adversarial effect of EvadeDroid. However, the number of AEs employed has an effect on the DREBIN's utility (i.e., the original performance of DREBIN). Table~\ref{table:performance_drebin_augmented_drebin} reveals that the addition of more AEs to the training set reduces the TPR of DREBIN. For instance, the TPR of DREBIN is reduced by $32.7\%$ compared to the standard training when $1769K$ malware samples in the training set are transformed into AEs. It is noteworthy that out of the $2K$ malware samples in the training set, EvadeDroid is capable of generating $1769$ AEs.




\end{document}